%% file: acl_latex.tex
\pdfoutput=1

\documentclass[11pt]{article}

\usepackage[]{latex/acl}

\usepackage{times}
\usepackage{latexsym}

\usepackage[T1]{fontenc}

\usepackage[utf8]{inputenc}

\usepackage{microtype}

\usepackage{inconsolata}

\input{latex/math_commands.tex}

\usepackage{hyperref}
\usepackage{url}

\usepackage{booktabs}
\usepackage{pifont}

\usepackage{multirow}
\usepackage{multicol}
\usepackage{longtable}
\usepackage{colortbl}
\usepackage[linewidth=0.8pt]{mdframed}

\usepackage[linesnumbered, ruled,vlined]{algorithm2e}
\usepackage{algorithmic}
\usepackage{graphicx}
\usepackage{wrapfig}
\usepackage{soul}

\usepackage{amsfonts}       
\usepackage{nicefrac}       
\usepackage{makecell}
\usepackage{subcaption}
\usepackage{parskip}
\usepackage{float}
\usepackage{array}
\usepackage{tabularx}
\usepackage{bm}
\usepackage{amsopn}
\usepackage{amsmath}
\usepackage{xcolor}
\usepackage{color}
\usepackage{multibib}
\usepackage{verbatim}
\usepackage{bbm}
\usepackage{bbding}
\usepackage{natbib}
\usepackage{arydshln}
\definecolor{OliveGreen}{rgb}{0,0.6,0}

\newcommand{\hs}[1]{\textcolor{magenta}{{#1}}}

\newcommand{\nop}[1]{}

\definecolor{myred}{HTML}{F8CCCC}
\definecolor{mygreen}{HTML}{E0ECD4}

\title{AttributionBench: How Hard is Automatic Attribution Evaluation?}

\author{Yifei Li \quad Xiang Yue \quad Zeyi Liao \quad Huan Sun \\
The Ohio State University \\
\text{{\texttt{\{li.14042, yue.149, liao.629, sun.397\}@osu.edu}}} \\
}

\begin{document}
\maketitle

\input{latex/abstract}
\input{latex/introduction}
\input{latex/benchmark}
\input{latex/experimental_setup}
\input{latex/results}
\input{latex/error_analysis}
\input{latex/related_work}
\input{latex/conclusion}
\input{latex/Limitations}
\input{latex/Ethics_Statement}

\bibliography{anthology,custom}

\input{latex/appendix}

\end{document}

%% file: latex/math_commands.tex
\usepackage{amsmath,amsfonts,bm}

\def\eqref#1{equation~\ref{#1}}

\def\1{\bm{1}}

\DeclareMathAlphabet{\mathsfit}{\encodingdefault}{\sfdefault}{m}{sl}
\SetMathAlphabet{\mathsfit}{bold}{\encodingdefault}{\sfdefault}{bx}{n}



%% file: latex/abstract.tex
\begin{abstract}

    Modern generative search engines enhance the reliability of large language model (LLM) responses by providing cited evidence. However, evaluating the answer's attribution, i.e., whether every claim within the generated responses is fully supported by its cited evidence, remains an open problem. 
    This verification, traditionally dependent on costly human evaluation, underscores the urgent need for automatic attribution evaluation methods. To bridge the gap in the absence of standardized benchmarks for these methods, we present \texttt{AttributionBench}, a comprehensive benchmark compiled from various existing attribution datasets. Our extensive experiments on \texttt{AttributionBench} reveal the challenges of automatic attribution evaluation, even for state-of-the-art LLMs. Specifically, our findings show that even a fine-tuned GPT-3.5
    only achieves around 80\% macro-F1 under a binary classification formulation.
    A detailed analysis of more than 300 error cases indicates that a majority of failures stem from the model's inability to process nuanced information, and the discrepancy between the information the model has access to and that human annotators do.
    \footnote{Our code and datasets are available at: \href{https://github.com/OSU-NLP-Group/AttributionBench}{https://github.com/OSU-NLP-Group/AttributionBench}}
\end{abstract}

%% file: latex/introduction.tex
\section{Introduction}
\label{sec:intro}

The advent of large language models (LLMs) has revolutionized the field of information retrieval and grounded text generation \citep{gpt3, wei2021finetuned, chowdhery2023palm, ouyang2022training, peng2023instruction}, leading to the development of advanced generative search engines like Bing Chat, Google Bard, and perplexity.ai. These platforms excel in generating search results in a natural language format, along with references to the source web pages as evidence to support the truthfulness of the generated response \citep{dziri-etal-2022-origin, 10.1145/3571730}. However, whether the evidence supports the generated responses or not, also known as the \textit{attribution} of the response to the evidence, remains an open problem.

\input{figures/attributionbench}

Many efforts \citep{liu-etal-2023-evaluating, kamalloo2023hagrid, malaviya2023expertqa} 
 have recently been made to conduct human evaluation to examine the performance of attribution of various advanced systems like Bing Chat and GPT-4. It turns out such systems often produce attribution errors, making them less faithful and trustworthy for practical use. However, human evaluation is expensive and time-consuming.\footnote{According to \citet{liu-etal-2023-evaluating}, the cost of annotation is around \$15 per hour, yielding around 15 data examples.} Therefore, there is a pressing need for efficient and effective methodologies to automatically assess attribution and detect possible errors.

Towards this end, there have been efforts made to build automatic attribution evaluation models. \citet{rashkin2021measuring} proposed the framework of \textit{Attributable to Identified Sources (AIS)},
i.e., evaluating whether model-generated responses can be verified against given references. Under this framework, several approaches were proposed to build the automatic evaluator, including directly utilizing natural language inference (NLI)
models \citep{gao-etal-2023-rarr, bohnet2022attributed}, directly prompting LLMs like GPT-3.5 and GPT-4 \citep{yue-etal-2023-automatic}, and fine-tuning smaller
LMs like FLAN-T5 \citep{chung2022scaling} with repurposed data from
related
tasks like NLI, fact-checking, and summarization \citep{yue-etal-2023-automatic}. 
However, these studies define the attribution evaluation task differently (i.e., using different numbers and types of classification labels) and choose different datasets to train and evaluate models,
making it impossible to directly compare their performance and derive significant insights into the challenges of attribution evaluation and how to address them in future work.


To this end, we take the first step to present a
systematic benchmark, \texttt{AttributionBench}, for training and evaluating cutting-edge automatic attribution evaluators.
Specifically, we meticulously sample data from 7 different datasets \citep{malaviya2023expertqa, liu-etal-2023-evaluating, bohnet2022attributed, chen2023understanding, dziri-etal-2022-evaluating, yue-etal-2023-automatic, kamalloo2023hagrid}
 that cover different domains of questions and diverse responses and evidence. We unify them into a binary classification format with a label-balanced setting for fair comparison\footnote{There can be more challenging problem formulations; but we will show that even if in this relatively simple formulation, there is still a large room for models to improve.}. We compile them into a training set and two test sets for in-distribution (ID) and out-of-distribution (OOD) evaluation, respectively.
Table \ref{tab:data_statistics} shows the statistics of our benchmark.


We conduct extensive experiments and analysis on our proposed benchmark. Surprisingly, we find that even fine-tuned GPT-3.5 can only get
arount 80\%
macro-F1 score
under both ID and OOD settings,
which is far away from practical use. To better understand the challenges of this task, we carefully labeled over 300 error cases from GPT-3.5 under chain-of-thought (CoT) prompting \citep{wei2022chain}, which generate rationales for the model's prediction that can reveal reasons for an error.
We find that 1) over 66\% errors are caused by the model's insensitivity to fine-grained information (i.e., ignoring or overlooking details like facts, events, numbers, or lack of necessary summarization and inference from the evidence, tasks that humans naturally perform),
and 2) about 26.8\% of the errors are caused by the mismatch between information accessible to the model and that accessible to human annotators (see examples in Figure \ref{fig:attributionbench}).
This is because the references given to the models are manually extracted from web pages by human annotators based on what they think is useful. However, when making the judgment, human annotators may be inherently affected by additional content since they've seen the whole webpage.
On the other hand, sometimes the supported evidence spreads across the entire webpage, making it impossible to locate a short snippet to serve as the evidence.
These findings provide valuable insights into 1) the challenges for automatic attribution evaluation, 2) the challenges for humans to evaluate attribution, and 3) how to develop stronger automatic attribution evaluation models.

We summarize our contributions as follows:

\noindent{\textbf{Benchmark.} We propose \texttt{AttributionBench}, a comprehensive benchmark with a unified formulation for attribution evaluation,
which will enable the community to compare different methods fairly and track the progress on this important task.}

\noindent{\textbf{Methods}.} We conduct comprehensive experiments and show that existing cutting-edge LLMs like GPT-4 and fine-tuned GPT-3.5 
still cannot perform well on this task.

\noindent{\textbf{Analysis.}} Through a series of in-depth error analyses, we show insights into why automatic attribution evaluation is difficult and potential future work.

    

\input{tables/data_statistics}

%% file: figures/attributionbench.tex
\begin{figure}[t]
    \centering
    \begin{minipage}{.5\textwidth}
        \includegraphics[width=\textwidth]{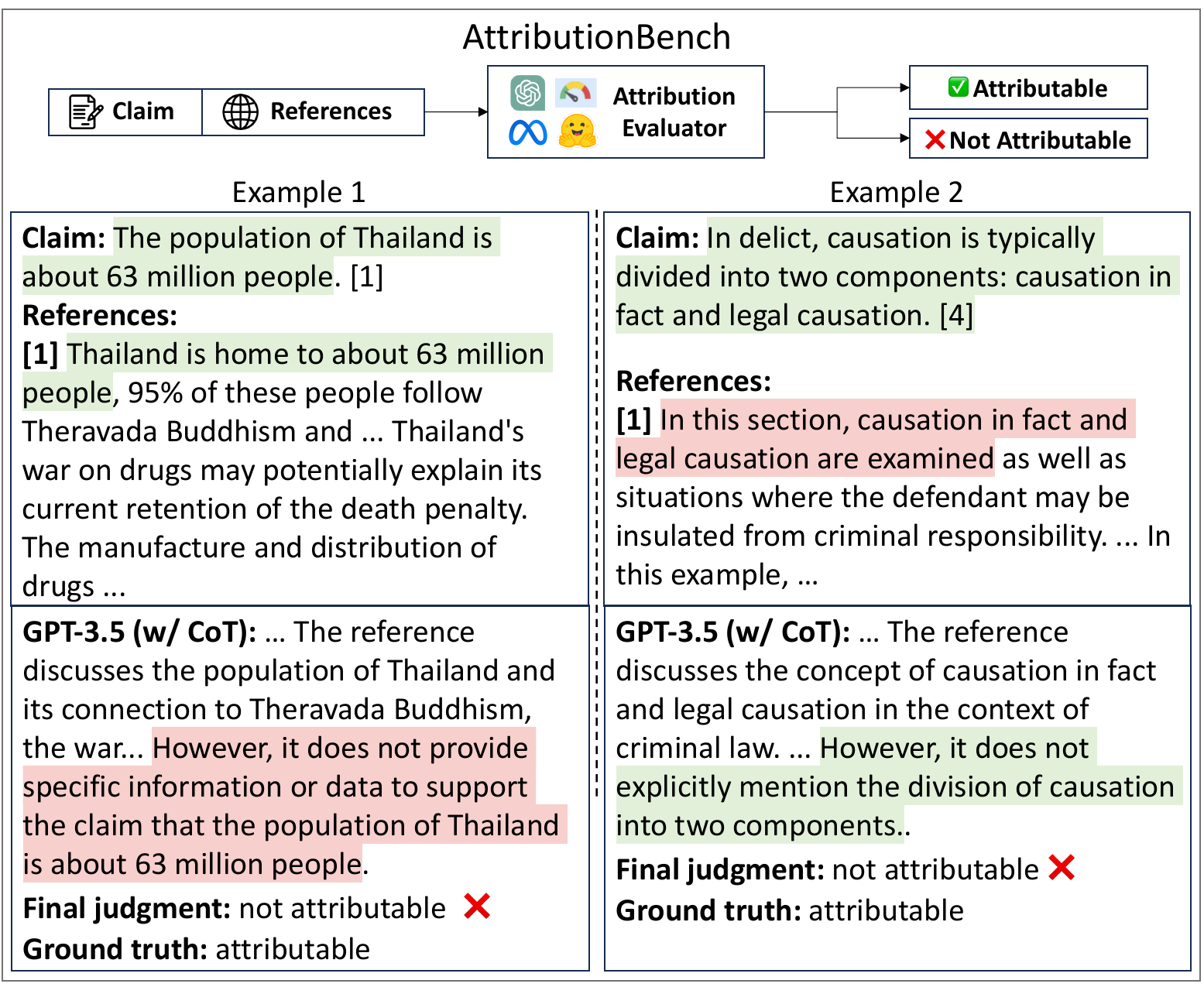}
        \caption{
        The illustration of the attribution evaluation task and two typical error examples from \textbf{AttributionBench} generated by GPT-3.5 (w/ CoT).
        The references are usually manually extracted from webpages by human annotators based on what they think is useful. Left: fine-grained information insensitivity (i.e., the model disregarded or overlooked nuanced details in either the claim or the references, as well as failing to do necessary summarization or inference from the given references, tasks that humans naturally perform). Right: human-model
        accessible
        information mismatch
        (i.e., human annotators can see the whole webpage while the model
        is only given
        the extracted evidence, leading to different judgments.)
        }
        \label{fig:attributionbench}
    \end{minipage}
\end{figure}

%% file: tables/data_statistics.tex
\begin{table}[t]
\renewcommand\arraystretch{1.15}
\centering
\resizebox{0.92\linewidth}{!}{
\centering
\begin{tabular}{p{0.25\linewidth}<{\raggedright}p{0.6\linewidth}<{\raggedright}p{0.12\linewidth}<{\raggedright}p{0.08\linewidth}<{\raggedright}p{0.08\linewidth}<{\raggedright}}

\toprule[1pt]
\multirow{1}{0.25\linewidth}{\textbf{Dataset}} & \multirow{1}{0.6\linewidth}{\makecell[Xl]{\textbf{Question Sources}}} & \multirow{1}{0.12\linewidth}{\textbf{\#Train}} & \multirow{1}{0.08\linewidth}{\textbf{\#Dev}} & \multirow{1}{0.08\linewidth}{\textbf{\#Test}} \\
\cmidrule{1-5}
\multirow{1}{0.25\linewidth}{ExpertQA} & \multirow{1}{0.6\linewidth}{\makecell[Xl]{Curated by domain experts}} & \multirow{1}{0.12\linewidth}{4442} & \multirow{1}{0.08\linewidth}{470} & \multirow{1}{0.08\linewidth}{612} \\
\cdashline{1-5}[1pt/1.5pt]
\multirow{3}{0.25\linewidth}{Stanford-GenSearch} & \multirow{3}{0.76\linewidth}{\makecell[Xl]{AllSouls, davinci-debate,\\ELI5,WikiHow-Keywords,\\NaturalQuestions}} & \multirow{3}{0.08\linewidth}{4784} & \multirow{3}{0.08\linewidth}{544} & \multirow{3}{0.08\linewidth}{600} \\
& & & & \\
& & & & \\
\cdashline{1-5}[1pt/1.5pt]
AttributedQA & NaturalQuestions & 2000 & 74 & 230 \\
\cdashline{1-5}[1pt/1.5pt]
LFQA & ELI5 & 2096 & 110 & 168 \\
\cdashline{1-5}[1pt/1.5pt]
\textbf{ID-Total} & & 13322 & 1198 & 1610 \\
\midrule[1pt]
\multirow{2}{0.25\linewidth}{BEGIN} & \multirow{2}{0.76\linewidth}{\makecell[Xl]{Wizard-of-Wikipedia,\\CMU-DoG,TopicalChat}} & \multirow{2}{0.08\linewidth}{\text{-}} & \multirow{2}{0.08\linewidth}{\text{-}} & \multirow{2}{0.08\linewidth}{436} \\
& & & & \\
\cdashline{1-5}[1pt/1.5pt]
HAGRID & MIRACL & - & - & 1088 \\
\cdashline{1-5}[1pt/1.5pt]
\multirow{2}{0.25\linewidth}{AttrEval-GenSearch} & \multirow{2}{0.76\linewidth}{Curated by human annotators} & \multirow{2}{0.08\linewidth}{\text{-}} & \multirow{2}{0.08\linewidth}{\text{-}} & \multirow{2}{0.08\linewidth}{162} \\
 & & & & \\
 \cdashline{1-5}[1pt/1.5pt]
\textbf{OOD-Total} & & - & - & 1686 \\
\bottomrule[1pt]
\end{tabular}
}
\caption{The datasets used in \textbf{AttributionBench}.
They contain a wide range of diverse questions. ``ID'' and ``OOD'' denote "in-distribution" and "out-of-distribution", respectively. We make all subsets label-balanced
so that the evaluation can fairly show the performance of each class. More details are listed in Appendix Table \ref{tab:appendix_data_statistics_complete}.
}
\label{tab:data_statistics}
\end{table}

%% file: latex/benchmark.tex
\section{\texttt{AttributionBench}}
\label{sec:benchmark}

In this section, we introduce our task formulation and data collection pipeline for constructing \texttt{AttributionBench}.

\subsection{Task Formulation}

We define the task of claim-level attribution evaluation as follows: Given a natural language query $Q$ and a response $\mathcal{R} = \{c_1, c_2, ..., c_n\}$ generated by LMs, comprising several claims $c_i$.
There is also an evidence set $E=\{e_1, e_2, ..., e_n\}$ along with each claim
The task is to judge whether a claim is supported by its accompanying evidence.
The claim-level evaluation model, denoted as a classifier $f(\cdot)$, is designed to take $\langle (Q), c_i, (\mathcal{R}), concat(e_i)\rangle$ as input. Here, the inclusion of the question $Q$ and the response $R$ in the input is optional since the
objective is to assess whether the evidence $\{e_i\}$ substantiates the claim $c_i$, and hence we use $(Q)$ and $(\mathcal{R})$; $concat(e_i)$ represents the concatenation of all corresponding evidences $e_{i_j}$ for claim $c_i$ Other information fields, if included, serve as supplementary data aiding the attribution evaluation task.


In previous work, the label space was defined differently, including binary classification (``Y'', ``N'') \citep{bohnet2022attributed},
three-way classification (``Support'', ``Partially Support'', ``No Support'') \citep{liu-etal-2023-evaluating} or (``Attributable'', ``Contradictory'', ``Extrapolatory'') \citep{yue-etal-2023-automatic}
To simplify and unify the task format, we formulate the task as a binary classification task, with the labels being (``Attributable'', ``Not Attributable'').


\subsection{Data Collection}

\paragraph{Datasets.} To make our benchmark more comprehensive, we collect data from several recently proposed attribution datasets, including ExpertQA \citep{malaviya2023expertqa}, Stanford-GenSearch \citep{liu-etal-2023-evaluating}, AttributedQA \citep{bohnet2022attributed}, LFQA \citep{chen2023understanding}, BEGIN \cite{dziri-etal-2022-evaluating}, AttrEval-GenSearch \citep{yue-etal-2023-automatic}, and HAGRID \citep{kamalloo2023hagrid}. 
These datasets contain different query sources, various domains, and are of varying difficulties.
We list the basic statistics in Table \ref{tab:data_statistics}, and more detailed statistics in Appendix Table \ref{tab:appendix_data_statistics_complete}.

\paragraph{Label-Balanced Setting.}
We select data samples from each contributing dataset and adopt a label-balanced setting in train, validation, and test set to reduce the effect of unbalanced labels on training and reliable evaluation of each label.
Such a balanced training set eliminates the potential negative impact brought by imbalanced labels,
and a balanced test set treats two classes equally and enhances the representativity of the F1 score.

\paragraph{In-Distribution \& Out-of-Distribution Test Sets.}
To measure both the effectiveness of fine-tuning and generalizability, we present both in-distribution and out-of-distribution test sets. We mainly consider 1) data scale and 2) there should be no question overlap between in-distribution (ID) and out-of-distribution (OOD) test sets, and pick ExpertQA, Stanford-GenSearch, AttributedQA, and LFQA as in-domain tasks, leaving BEGIN, AttrEval-GenSearch, and HAGRID as out-of-domain tasks.




%% file: latex/experimental_setup.tex
\section{Exerimental Setup}
\label{sec:experimental_setup}

\input{tables/id_results}



\paragraph{Datasets.} We fine-tune models with the training set of \texttt{AttributionBench} and use the ID and OOD test set of \texttt{AttributionBench} for evaluation.

\paragraph{Input Fields.} 
By default,
we only include $c$ and $E$
as input, ignoring $Q$ and $R$, as this setting shows the best performance preliminarily on GPT-3.5 (section \ref{sec:input_fields}.

\paragraph{Compared Models.}
We include multiple types of models in our main experiments. (1) decoder-only models: Llama-2 (7B) \citep{touvron2023llama}, AttrScore-Alpaca (7B) \citep{yue-etal-2023-automatic}, GPT-3.5 \citep{openai2023chatgpt} and GPT-4 \citep{achiam2023gpt}. (2) encoder-decoder models: FLAN-T5 (770M, 3B, 11B) \citep{chung2022scaling}, AttrScore-FLAN-T5 (3B) \citep{yue-etal-2023-automatic}, T5-XXL-TRUE (11B) \citep{honovich-etal-2022-true-evaluating}, and FLAN-UL2 (20B) \citep{tay2022ul2}. (3) encoder-only models: \texttt{roberta-large-mnli} (340M) \citep{liu2019roberta}. Note that, \texttt{roberta-large-mnli} and T5-XXL-TRUE are fine-tuned on natural language inference (NLI) data, and AttrScore-FLAN-T5 (3B) and AttrScore-Alpaca (7B) are fine-tuned on data repurposed from similar tasks including NLI, fact-checking, QA, and summarization.
Each model (except \text{GPT-4}) is tested via both zero-shot prompting and fine-tuning. The implementation details are listed in Appendix \ref{sec:appendix_implementation_details}.

%% file: tables/id_results.tex
\begin{table*}[ht]
\resizebox{0.99\linewidth}{!}{
\centering
\begin{tabular}{cl|ccc|ccc|ccc|ccc|c}

\toprule[1pt]
\multirow{2}{*}{\textbf{Setting}} & \multirow{2}{*}{\textbf{Model (Size)}} & \multicolumn{3}{c|}{\textbf{ExpertQA}} & \multicolumn{3}{c|}{\textbf{Stanford-GenSearch}} & \multicolumn{3}{c|}{\textbf{AttributedQA}} & \multicolumn{3}{c|}{\textbf{LFQA}} & \textbf{ID-Avg.} \\ \cmidrule{3-15}
 & & F1 $\uparrow$ & FP $\downarrow$ & FN $\downarrow$ & F1 $\uparrow$ & FP $\downarrow$ & FN $\downarrow$ & F1 $\uparrow$ & FP $\downarrow$ & FN $\downarrow$ & F1 $\uparrow$ & FP $\downarrow$ & FN $\downarrow$ & F1 $\uparrow$ \\ \midrule[1pt]
\multirow{11}{*}{Zero-shot} & FLAN-T5 (770M) & 38.2 & 1.3 & 47.4 & 73.5 & 15.0 & 11.5 & 80.4 & 12.2 & 7.4 & 37.2 & 0.0 & 48.2 & 57.3 \\
 & FLAN-T5 (3B) & 55.6 & 15.8 & 27.9 & \textbf{74.0} & 17.2 & 8.7 & 79.8 & 15.2 & 4.8 & 75.3 & 6.5 & 17.9 & 71.2 \\
 & AttrScore-FLAN-T5 (3B) & 55.7 & 32.4 & 9.6 & 64.6 & 27.3 & 6.5 & 80.5 & 16.5 & 2.6 & 71.4 & 21.4 & 6.5 & 68.1 \\
 & FLAN-T5 (11B) & 52.0 & 36.4 & 7.5 & 59.2 & 32.7 & 5.0 & 78.6 & 18.3 & 2.6 & 79.8 & 10.1 & 10.1 & 67.4 \\
 & T5-XXL-TRUE (11B) & 54.5 & 17.8 & 27.3 & 68.5 & 16.2 & 15.3 & \textbf{85.2} & 7.8 & 7.0 & \textbf{80.4} & 1.2 & 17.9 & 72.2 \\
 & FLAN-UL2 (20B) & \textbf{59.4} & 22.5 & 18.0 & 72.5 & 19.2 & 8.0 & 82.5 & 13.0 & 4.3 & 80.1 & 4.2 & 15.5 & \textbf{73.6} \\
\cmidrule{2-15}
 & AttrScore-Alpaca (7B) & 47.4 & 11.1 & 37.7 & 68.6 & 21.2 & 9.8 & 79.0 & 14.8 & 6.1 & 68.7 & 10.1 & 20.8 & 65.9 \\
 & GPT-3.5 (w/o CoT) & 55.3 & 30.4 & 12.1 & 62.0 & 30.5 & 3.8 & 74.7 & 20.9 & 3.5 & 72.6 & 22.0 & 4.2 & 66.2 \\
 & GPT-3.5 (w/ CoT) & \textbf{60.4} & 23.0 & 16.2 & 66.1 & 25.5 & 7.2 & 78.9 & 14.3 & 6.5 & 73.4 & 19.6 & 6.5 & 69.7 \\
 & GPT-4 (w/o CoT) & 56.5 & 32.8 & 8.0 & 59.8 & 33.2 & 3.5 & 81.0 & 15.7 & 3.0 & 71.6 & 23.2 & 4.2 & 67.2 \\
 & GPT-4 (w/ CoT) & 59.2 & 26.3 & 13.9 & \textbf{71.7} & 19.5 & 8.5 & \textbf{82.2} & 10.0 & 7.8 & \textbf{80.2} & 14.9 & 4.8 & \textbf{73.3} \\
\midrule[1pt]
\multirow{11}{*}{Fine-tuned} & Roberta-large-mnli (330M) & 52.0 & 13.1 & 33.0 & 64.7 & 14.0 & 21.2 & 71.5 & 10.0 & 18.3 & 68.0 & 24.4 & 6.5 & 64.1 \\
\cmidrule{2-15}
 & FLAN-T5 (770M) & 55.0 & 32.7 & 10.0 & 75.2 & 16.0 & 8.7 & 81.6 & 13.5 & 4.8 & 83.3 & 9.5 & 7.1 & 73.8 \\
 & FLAN-T5 (3B) & 54.9 & 33.8 & 8.3 & 79.8 & 10.7 & 9.5 & 82.1 & 12.2 & 5.7 & 89.3 & 6.0 & 4.8 & 76.5 \\
 & AttrScore-FLAN-T5 (3B) & 56.8 & 30.2 & 11.4 & 81.0 & 10.0 & 9.0 & 82.5 & 11.7 & 5.7 & \textbf{90.5} & 4.2 & 5.4 & 77.7 \\
 & FLAN-T5 (11B) & \textbf{61.8} & 21.9 & 16.2 & 81.7 & 9.5 & 8.8 & 83.4 & 11.7 & 4.8 & 86.9 & 6.0 & 7.1 & \textbf{78.5} \\
 & T5-XXL-TRUE (11B) & 57.1 & 30.1 & 11.3 & \textbf{81.8} & 9.5 & 8.7 & 83.4 & 12.6 & 3.9 & 86.3 & 4.8 & 8.9 & 77.2 \\
 & FLAN-UL2 (20B) & 61.6 & 24.8 & 13.1 & \textbf{81.8} & 8.2 & 10.0 & \textbf{84.3} & 10.0 & 5.7 & 86.3 & 7.1 & 6.5 & \textbf{78.5} \\
\cmidrule{2-15}
 & Llama-2 (7B) & 60.5 & 17.2 & 22.2 & 80.2 & 10.5 & 9.3 & 79.9 & 13.5 & 6.5 & \textbf{85.6} & 3.0 & 11.3 & 76.6 \\
 & AttrScore-Alpaca (7B) & \textbf{61.3} & 16.3 & 22.2 & \textbf{82.8} & 8.2 & 9.0 & 80.4 & 11.3 & 8.3 & 84.5 & 6.5 & 8.9 & 77.3 \\
 & GPT-3.5 (w/o CoT) & 61.1 & 18.8 & 19.9 & 82.0 & 6.7 & 11.3 & \textbf{83.9} & 8.7 & 7.4 & 83.9 & 6.5 & 9.5 & \textbf{77.7} \\
\midrule[1pt]
 & Avg. Gain$^{*}$ & 6.4 & 6.0 & -9.9 & 12.4 & -11.9 & 0.4 & 2.1 & -2.6 & 0.6 & 15.6 & -2.8 & -10.7 & 9.0 \\
\bottomrule[1pt]
\end{tabular}
}
\caption{The baseline performance (macro-F1 score) with different models on the 4 ID test sets of \texttt{AttributionBench}. The best performance within each setting is marked in \textbf{bold}. FP: false positive
percentage (the ground truth is ``not attributable'' but the model predicts ``attributable''). FN: false negative percentage (
vice versa
).
ID-Avg.: the average score of F1 on 4 ID test sets. ``$\uparrow$'' and ``$\downarrow$'' represent higher and lower is better, respectively.
$^*$: The ``Avg. Gain'' is calculated by averaging the performance gain of each model that is both zero-shot prompted and fine-tuned) after being fine-tuned compared to being zero-shot prompted (If a model is only in one setting, its gain isn't included in the average).
The results demonstrate the effectiveness of fine-tuning on the training set of \texttt{AttributionBench} on ID evaluation, but there is still a large room for improvement.
}
\label{tab:in-domain-results}
\end{table*}

%% file: latex/results.tex
\section{Result Analysis}
\label{sec:results}

\subsection{Main \nop{\hs{Main}} Results}
\input{tables/ood_results}
Tables \ref{tab:in-domain-results} and \ref{tab:out-of-domain-results} show the overall results of different models on ID and OOD test sets.
Our primary findings are as follows:


\paragraph{Fine-tuning on NLI-related data is beneficial to attribution evaluation.}
T5-XXL-TRUE and AttrScore-Flan-T5 (3B) are fine-tuned on data including NLI and achieve strong performance on both ID and OOD sets. Specifically, both models outperform fine-tuned GPT-3.5 for the average F1 score on OOD sets and achieve comparable performance with fine-tuned GPT-3.5 on OOD sets. 
Also, through further fine-tuning on \texttt{AttributionBench}, AttrScore-Flan-T5 (3B) outperforms the original Flan-T5 (3B) on 4 ID and 2 OOD sets.
For T5-XXL-TRUE, it surpasses FLAN-T5 (11B), a model with the same amount of parameters and additionally trained on instruction data, on 2 ID and 2 OOD sets.
These results indicate the effectiveness of training on NLI data.
A possible reason is that the attribution evaluation task shares a similar ability with solving NLI tasks, as both of them need the model to judge the entailment (being supported) of the hypothesis (claim) against the premise (evidence).

\paragraph{Automatic attribution evaluation is challenging under zero-shot setting.}
Attribution evaluation is a complex task because it not only requires a comprehensive understanding, necessary summarization, and inference of the referenced information but also requires a fine-grained comparison with the content of the claim.
Moreover, this requirement criteria may vary considerably across different datasets and testing conditions.
In this challenging landscape, models often demonstrate suboptimal performance in zero-shot scenarios.
For tasks with shorter claims and evidence like AttribtuedQA and AttrEval-GenSearch, the performance of GPT-3.5 and GPT-4 are over 80\%, while for tasks with longer evidence (like Stanford-GenSearch and HAGRID), the zero-shot performance for models is around 60\%\textasciitilde70\%, which is relatively low. For the challenging task ExpertQA, which consists of domain-specific challenging questions, GPT-3.5, GPT-4 and those NLI-tuned models can only achieve under \textasciitilde60\% performance.

\paragraph{Fine-tuning on \texttt{AttributionBench} benefits both ID and OOD evaluation.}
Despite the task's challenge, the performance can be enhanced through fine-tuning on the training set of \texttt{AttributionBench}. On average among all models, fine-tuning can improve $9.0\%$ and $4.6\%$ on 4 in-domain tasks and 3 out-of-domain tasks, respectively. Additionally, with training on only 13k examples, the fine-tuned FLAN-T5 (770M) model even surprisingly outperforms GPT-3.5 (w/ CoT) on both ID and OOD sets, indicating the effectiveness of fine-tuning on \texttt{AttributionBench}.
Furthermore, almost all models obtain performance gain on 3 OOD sets via fine-tuning, indicating that the models did not just overfit the ID data, but also gained generalizability to solve this task on OOD examples.

\paragraph{Simply switching stronger models cannot significantly improve the performance.}
First, under zero-shot setting, GPT-3.5 and GPT-4 show their expertise in dealing with difficult tasks like ExpertQA, a dataset containing hard domain-specific responses and evidence.
Nevertheless, on more other tasks like AttributedQA and LFQA, they still underperform smaller models like T5-XXL-TRUE and FLAN-UL2.
Under fine-tuned setting, for the 4 ID sets, GPT-3.5 shows competing performance but still underperforms smaller models including FLAN-T5 (11B) and Flan-UL2 (20B).
For the 3 OOD sets, the performance of GPT-3.5 on AttrEval-GenSearch and HAGRID is still lower than many models including T5-XXL-TRUE, AttrScore-Flan-T5 (3B), and most surprisingly, Flan-T5 (3B) and Flan-T5 (770M).
Although it is hard for one model to do well among all datasets with various data distributions and different classification criteria, both zero-shot and fine-tuning results indicate that simply scaling up and fine-tuning models on ID data isn't the solution to producing a strong attribution evaluator model for practical use.

\subsection{The Impact of Input Fields}
\label{sec:input_fields}
\input{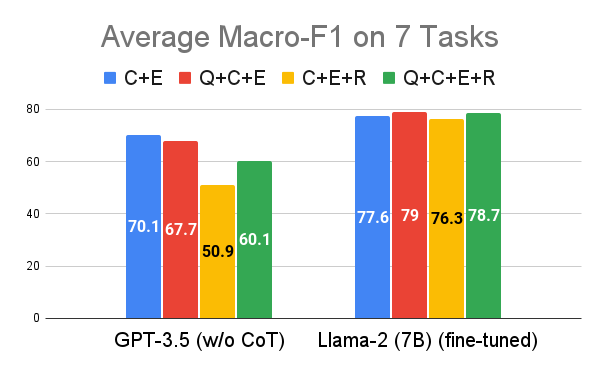}
Figure \ref{fig:input_fields} illustrates the impact of incorporating both questions and responses within the model input.
The result reveals that \textbf{merely appending the questions and responses to the input, which human annotators might have access to, is not the key to solving this task}.
Despite providing more contextual information, both adding $Q$ and adding $R$ do not improve the performance. By looking at error cases, we find that although questions introduce essential elements that could influence the model's judgment, they could also potentially mislead the model by confusing the attribution task with an assessment of ``answer usefulness'' instead of attribution. Specifically, in instances where the evidence substantiates the claim but fails to precisely address the question, the model erroneously dismisses it as ``not attributable''. This issue is similar when adding the response field, as such amplification of context could exacerbate the model's confusion regarding the specific details within the input text.


\subsection{The Impact of Prompts}
\input{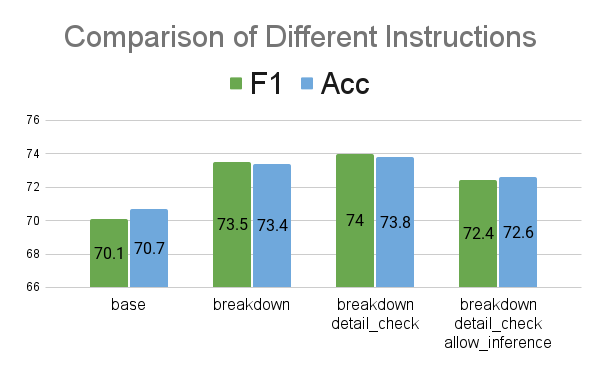}

It is well known that the choice of prompts
can lead to different performances.
Here we investigate the impact of instructions using GPT-3.5 with CoT.
Figure \ref{fig:impact_of_instructions} shows that \textbf{prompt engineering only brings limited gain for GPT-3.5 over 7 test sets}.
Using the base prompts,
we observe that there are more false positive cases than false negative ones, which means the model
tends to be positive about the attribution and predict a data sample to be attributable.
Therefore, we carefully constructed several prompts: (1) breakdown and verify step-by-step, (2)
on top of (1),
force the model to be more strict and check every factual detail within the claim,
and (3) based on (1) and (2), besides carefully checking details, the model is also required to do necessary inference and summarization to comprehensively understand the semantics of both the claim and the evidence.
 The detailed prompts we used are listed in Appendix Table \ref{tab:appendix_prompts}. The results show that although prompt engineering can bring some marginal gain in accuracy and F1-score, it is essentially changing the classification interval and making trade-offs between false positive and false negative cases.

%% file: tables/ood_results.tex
\begin{table*}[ht]
\centering
\resizebox{0.92\linewidth}{!}{
\begin{tabular}{cl|ccc|ccc|ccc|c}

\toprule[1pt]
\multirow{2}{*}{\textbf{Setting}} & \multirow{2}{*}{\textbf{Model (Size)}} & \multicolumn{3}{c|}{\textbf{BEGIN}} & \multicolumn{3}{c|}{\textbf{AttrEval-GenSearch}} & \multicolumn{3}{c|}{\textbf{HAGRID}} & \textbf{OOD-Avg.} \\ \cmidrule{3-12}
 & & F1 $\uparrow$ & FP $\downarrow$ & FN $\downarrow$ & F1 $\uparrow$ & FP $\downarrow$ & FN $\downarrow$ & F1 $\uparrow$ & FP $\downarrow$ & FN $\downarrow$ & F1 $\uparrow$ \\ \midrule[1pt]
\multirow{11}{*}{Zero-shot} & FLAN-T5 (770M) & 79.6 & 9.2 & 11.2 & 80.8 & 6.2 & 13.0 & 75.9 & 13.1 & 10.9 & 78.8 \\
 & FLAN-T5 (3B) & 80.2 & 13.3 & 6.4 & 82.0 & 6.2 & 11.7 & \textbf{79.0} & 16.9 & 3.8 & 80.4 \\
 & AttrScore-FLAN-T5 (3B) & 78.9 & 17.7 & 3.0 & 76.3 & 16.7 & 6.8 & 68.6 & 26.9 & 2.6 & 74.6 \\
 & FLAN-T5 (11B) & 72.3 & 25 & 1.1 & 78.1 & 16.7 & 4.9 & 64.5 & 30.6 & 2.0 & 71.6 \\
 & T5-XXL-TRUE (11B) & \textbf{86.4} & 4.8 & 8.7 & 76.4 & 2.5 & 20.4 & 78.6 & 14.4 & 6.8 & 80.5 \\
 & Flan-UL2 (20B) & 82.2 & 13.1 & 4.6 & \textbf{87.7} & 5.6 & 6.8 & 73.9 & 21.4 & 3.9 & \textbf{81.3} \\
\cmidrule{2-12}
 & AttrScore-Alpaca (7B) & 75.9 & 20.4 & 3.0 & 82.1 & 6.8 & 11.1 & 73.9 & 19.9 & 5.6 & 77.3 \\
 & GPT-3.5 (w/o CoT) & \textbf{79.4} & 15.8 & 4.4 & 76.7 & 18.5 & 4.3 & 70.1 & 25.2 & 2.8 & 75.4 \\
 & GPT-3.5 (w/ CoT) & 77.6 & 14.9 & 7.3 & 82.1 & 11.1 & 6.8 & 74.0 & 19.7 & 5.1 & 77.9 \\
 & GPT-4 (w/o CoT) & 77.5 & 19.7 & 2.1 & \textbf{84.3} & 14.2 & 1.2 & 72.1 & 23.9 & 2.8 & 78.0 \\
 & GPT-4 (w/ CoT) & 77.5 & 18.3 & 3.7 & 83.3 & 8.0 & 8.6 & \textbf{75.9} & 18.5 & 5.2 & \textbf{78.9} \\
\midrule[1pt]
\multirow{11}{*}{Fine-tuned} & Roberta-large-mnli (330M) & 60.9 & 6.2 & 27.2 & 65.1 & 26.1 & 11.8 & 61.3 & 9.2 & 28.4 & 62.4 \\
\cmidrule{2-12}
 & FLAN-T5 (770M) & 79.6 & 9.2 & 11.2 & 80.8 & 6.2 & 13.0 & 75.9 & 13.1 & 10.9 & 78.8 \\
 & FLAN-T5 (3B) & 80.2 & 13.3 & 6.4 & 82.0 & 6.2 & 11.7 & \textbf{79.0} & 16.9 & 3.8 & 80.4 \\
 & AttrScore-FLAN-T5 (3B) & 90.8 & 6.7 & 2.5 & \textbf{85.1} & 4.3 & 10.5 & 76.6 & 19.0 & 3.9 & \textbf{84.2} \\
 & FLAN-T5 (11B) & 90.6 & 7.8 & 1.6 & \textbf{85.1} & 4.3 & 10.5 & 67.7 & 27.1 & 3.4 & 81.1 \\
 &T5-XXL-TRUE (11B) & \textbf{91.5} & 7.1 & 1.4 & 81.4 & 6.2 & 12.3 & 77.3 & 18.6 & 3.6 & 83.4 \\
 & Flan-UL2 (20B) & 90.1 & 6.9 & 3.0 & \textbf{85.1} & 4.9 & 9.9 & 67.8 & 26.6 & 4.0 & 81.0 \\
\cmidrule{2-12}
 & Llama-2 (7B) & 84.1 & 10.6 & 5.3 & 83.3 & 6.8 & 9.9 & 73.0 & 22.5 & 3.5 & 80.1 \\
 & AttrScore-Alpaca (7B) & 85.1 & 9.2 & 5.7 & 81.9 & 4.3 & 13.6 & 76.4 & 18.3 & 4.9 & 81.1 \\
 & GPT-3.5 (w/o CoT) & \textbf{86.8} & 11.0 & 2.1 & 81.3 & 4.9 & 13.6 & \textbf{77.6} & 16.9 & 5.1 & \textbf{81.9} \\
\midrule[1pt]
 & Avg. Gain $^{*}$ & 9.9 & -6.6 & -3.2 & 2.5 & -4.1 & 1.6 & 1.0 & 0.3 & -1.2 & 4.6 \\
\bottomrule[1pt]
\end{tabular}
}
\caption{The baseline performance (Macro-F1 score) with different models on the 3 OOD test sets of \texttt{AttributionBench}. The best performance within each setting is marked in \textbf{bold}.
FP: false positive percentage (the ground truth is ``not attributable'' but the model predicts ``attributable'').
FN: false negative percentage (vice versa).
OOD-Avg.: the average score of F1 on 3 OOD test sets.
``$\uparrow$'' and ``$\downarrow$'' represent higher and lower is better, respectively.
$^*$: The ``Avg. Gain'' is calculated by averaging the performance gain of each
model (that is both zero-shot prompted and fine-tuned) after being fine-tuned compared to being zero-shot prompted.
Fine-tuned results demonstrate fine-tuning on the training set of \texttt{AttributionBench} not only improves ID performance but also improves generalizability.}
\label{tab:out-of-domain-results}
\end{table*}

%% file: figures/input_fields.tex
\begin{figure}[t]
    \centering
    \begin{minipage}{.5\textwidth} 
        \centering
        \includegraphics[width=\textwidth]{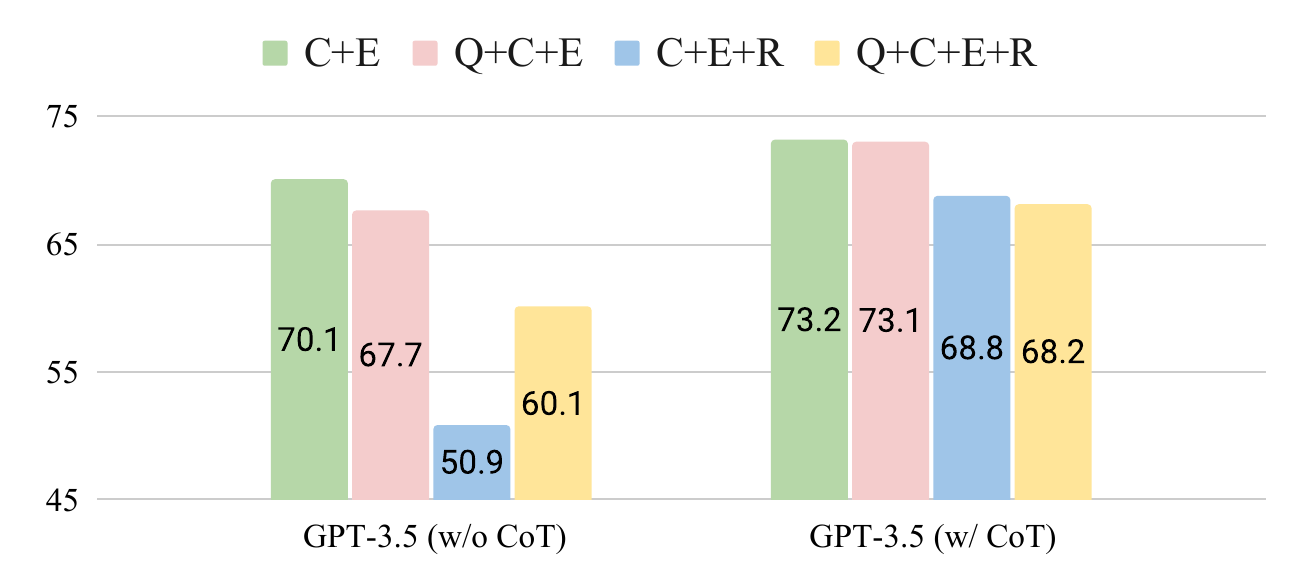} 
        \caption{The average macro-F1 score on 7 test sets of GPT-3.5 with different input fields. Q, C, E, R stands for question, claim, evidence, and response, respectively. Results show that despite involving additional information, adding input fields cannot boost or even harm the performance.
        }
        \label{fig:input_fields}
    \end{minipage}
\end{figure}

%% file: figures/impact_of_instructions.tex
    

\begin{figure}[t]
     \centering
     \begin{subfigure}[b]{0.5\textwidth}
         \centering
         \includegraphics[width=\textwidth]{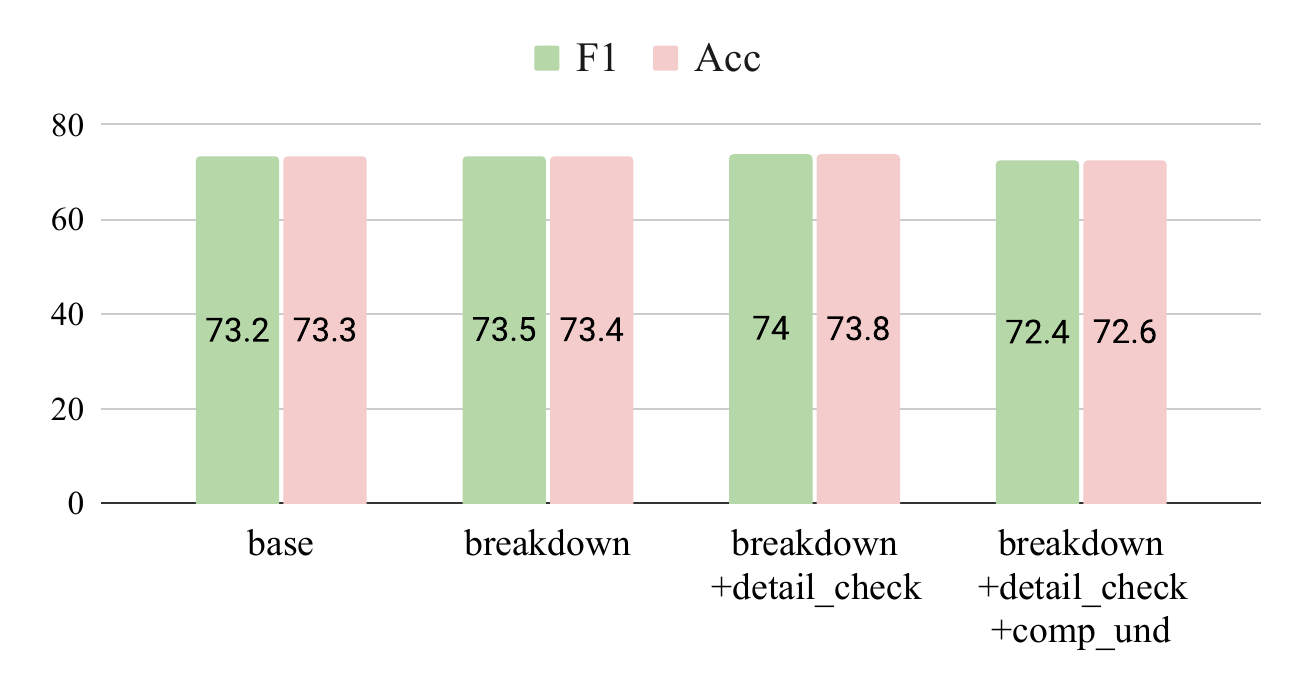}
         \caption{}
         \label{fig:impact_of_instructions_1}
     \end{subfigure}
     \hfill
     \begin{subfigure}[b]{0.5\textwidth}
         \centering
         \includegraphics[width=\textwidth]{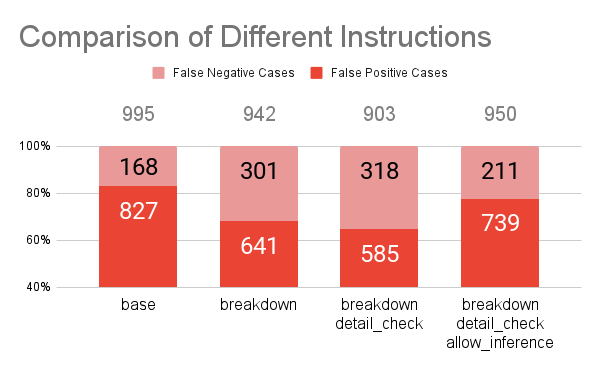}
         \caption{}
         \label{fig:impact_of_instructions_2}
     \end{subfigure}
        \caption{The performance of GPT-3.5 (w/ CoT) with several different prompts. Prompt engineering only brings limited gain over 7 test sets.
        Rather than bringing little gain in overall performance (Fig \ref{fig:impact_of_instructions_1}), adjusting prompts is actually changing the ratio of FP and FN cases (Fig \ref{fig:impact_of_instructions_2})
        ``comp\_und'' stands for ``comprehensive understanding''.}
        \label{fig:impact_of_instructions}
\end{figure}

%% file: latex/error_analysis.tex
\section{Error Analysis}
\label{sec:error_analysis}

After extensive baseline assessments and ablation studies, we sought to pinpoint the major challenges for the task of attribution evaluation.
We conduct a detailed qualitative error analysis using GPT-3.5 (w/ CoT) by randomly sampling 50 error cases (or all of them if fewer than 50) for each dataset, which results in 321 instances in total.
For ExpertQA, Stanford-GenSearch, BEGIN, and HAGRID, we pick 25 cases for both false positive and false negative examples and label all the error cases for AttributedQA, LFQA, and AttrEval-GenSearch.
We show representative examples in Table \ref{tab:data_statistics}.

\input{tables/error_case}

Our first observation is that \textbf{over $66\%$ error cases are caused by fine-grained information insensitivity.} This means when making a wrong judgment, the model is most likely to fail in comparing
fine-grained information within the claim and the evidence. Here, ``fine-grained information'' encompasses a wide array of specifics, from discrete data points such as numerical figures, dates, names, and locations, to more complex elements like particular events or logical connections.
Moreover, within this category of error, the model also makes different kinds of mistakes regarding its label error class, i.e., false positive or false negative. In the case of false positives, the model misses certain details within the claim 43\% of the time
, and in 41\% of the instances,
it made incorrect connections and inferences to indicate the claim being supported by the evidence, which is actually wrong.
Conversely, within the false negative category, the model predominantly misinterprets or neglects details in the evidence around 46\% of the time, and 39\% of them are because of lacking necessary inference and summarization to reach the attribution judgment, which could be very easy for humans.

\input{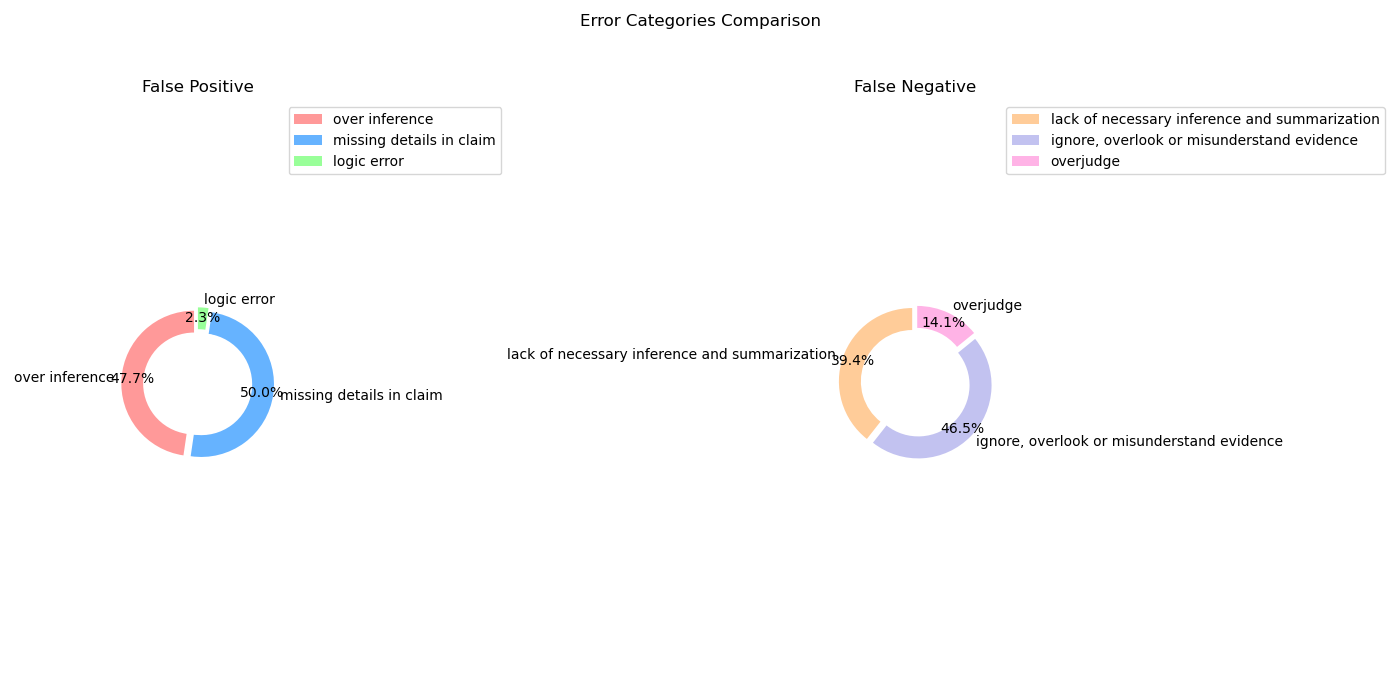}
\textbf{About $26.8\%$ errors are caused by the mismatch between information accessible to the model and that accessible to human annotators.}
This disparity brings into question the reliability of labels within current attribution evaluation datasets. Our analysis suggests that, in certain cases, determining the label based solely on the provided claim and evidence can be challenging due to insufficient information or ambiguous references (error type: need additional information/reference ambiguity).
At times, this can even lead to
the annotated label being wrong
(error type: label mismatch). However, this doesn't mean that nearly a quarter of the labels are wrong. According to the current annotation process \citep{liu-etal-2023-evaluating, malaviya2023expertqa}, human annotators usually assess claims sequentially, which allows for additional context to inform their judgment on subsequent claims after reviewing the preceding ones. Furthermore, while the models receive manually extracted snippets of evidence from web pages, human annotators have the advantage of
implicitly
considering the entire web page when making their judgments. This discrepancy between the annotator's access and the model's to information can lead to inconsistencies between the annotation and the prediction process. Consequently, even though many studies boast high concordance rates for their labels \citep{liu-etal-2023-evaluating, malaviya2023expertqa}, the true reliability of these metrics and the correctness of the labels warrant further investigation. 



\paragraph{Other error types.} The model also struggles with dealing with numerical/symbolic operators, misunderstanding the task definition and logical relation implied by labels, and sometimes fails due to the lack of implicit commonsense knowledge
Compared to the previous two error types, these errors can be more easily fixed by augmenting LMs with external tools, injecting necessary knowledge when prompting/inference, and making clearer instructions and in-context examples as guidance.


Our observations reveal the difficulty of the attribution evaluation task, leading to two potential directions for improvement: (1) For the model side, we should let the models not only be faithful, strongly conditioned
on the given evidence, and pay attention to fine-grained information, but also possess the necessary ability of summarizing and reasoning
over the evidence.
\nop{ This could be enhanced by methods like retrieval-augmented generation \citep{gao2023retrieval} or knowledge editing \citep{mishra2024fine};}
(2) For the data side, the given instruction and evidence
to the model
should be more aligned with what human annotators get. The detailed error ratios of each sub-dataset are listed in Appendix \ref{sec:appendix_error_analysis}.

%% file: tables/error_case.tex
\begin{table*}[t]
\renewcommand\arraystretch{1.2}
\small
\centering
\resizebox{0.92\linewidth}{!}{
\begin{tabular}{p{0.14\textwidth}<{\raggedright}p{0.86\textwidth}}
\toprule[1pt]
\textbf{Error Type (\%)} & \textbf{Example} \\
\midrule[1pt]

Fine-grained information insensitivity (66.0\%) &
\textbf{Claim:} He was once an important member of the Party but \colorbox{mygreen}{became a traitor}.

\textbf{Evidence:} He seems to have been a Party leader who \colorbox{mygreen}{fell out of favor with the regime}.

\textbf{Human-annotated Ground truth}: \colorbox{mygreen}{not attributable}

\textbf{GPT-3.5 (w/ CoT)}: The reference supports the claim as it indicates that the person was once an important member of the Party (Party leader) but then \colorbox{mygreen}{fell out of favor with the regime}, \colorbox{myred}{which aligns with the claim of} \colorbox{myred}{becoming a traitor}. Final judgment: \colorbox{myred}{attributable}. \\
\hline

Need additional information/reference ambiguity (15.6\%) & 
\textbf{Claim:} \colorbox{myred}{Kellie Pickler} won season 16 of Dancing with the Stars.

\textbf{Evidence:} Title: Dancing with the Stars (American season 16)$\backslash$n$\backslash$n\colorbox{mygreen}{Pickler} and her partner Derek Hough won the competition, marking his fourth of six wins and increasing his lead on the show as the professional dancer who has won the most mirror ball trophies.

\textbf{Human-annotated Ground truth}: \colorbox{mygreen}{not attributable}

\textbf{GPT-3.5 (w/ CoT)}: The reference supports the claim. \colorbox{mygreen}{Kellie Pickler} did indeed win season 16 of Dancing with the Stars, as mentioned in the reference. Final judgment: attributable. \\
\hline

Label issue (11.2\%) & 
\textbf{Claim:} Ireland last qualified for the World Cup in 2002, \colorbox{myred}{as highlighted in and}.

\textbf{Evidence:} By the time of their next potential appearance, in 2026, Ireland will not have been involved in the finals since they took La Roja to penalties in a campaign lit up by the goals of the talismanic Robbie Keane 24 years earlier.

\textbf{Human-annotated Ground Truth:} \colorbox{myred}{attributable}

\textbf{GPT-3.5 (w/ CoT):} The reference does not support the claim. The reference only mentions Ireland's potential appearance in the 2026 World Cup and their absence from the finals since 1994, but it does not specifically mention their last qualification in 2002. Therefore, the reference does not support the claim. Final judgment: \colorbox{mygreen}{not attributable}. \\
\hline





\end{tabular}
}
\caption{
Representative error examples from prompting GPT-3.5 (w/CoT) in \texttt{AttributionBench}.
\colorbox{mygreen}{Green background} indicates the correct spans and labels, and \colorbox{myred}{red background} indicates the wrong spans or labels.
Case 1: the model over inferred ``fell out of favor with the regime'' into ``became a traitor''. Case 2: The ``Pickler'' in the evidence is ambiguous, leading to the ``not attributable'' label by the human annotator while the model failed to capture it. Case 3: It's impossible to get an ``attributable'' conclusion according to the extracted evidence, whereas the original web page may support it.
}
\label{tab:error_examples}
\end{table*}

%% file: latex/related_work.tex
\section{Related Work}
\paragraph{Attributed LLMs.}
Generative LLMs often produce hallucinations, especially when encountering knowledge-intensive tasks, leading to the urgent need for attribution evaluation. Recently, many papers \citep{2302.07842v1, 2302.04761v1, li-etal-2023-api} have emerged to alleviate such issues by augmenting LLMs with external tools, such as retrievers \citep{pmlr-v119-guu20a, 2005.11401v4} and search engines \citep{nakano2021webgpt, komeili-etal-2022-internet}. Notably, some work explicitly equips generation with retrieved supporting citations or evidence as the attribution for the generated statement. For instance, GopherCite \citep{gophercite} puts the citation
after the generation as inline evidence. 
WebGPT \citep{nakano2021webgpt} finetunes GPT-3 \citep{gpt3} to simulate a person's behavior of searching online and finding relevant information. Dialogue systems like LaMDA \citep{lamda} and Sparrow \citep{sparrow} improve their safety and factual grounding by fine-tuning with human-annotated data and enabling the model to seek help from external resources. 


\paragraph{Attribution Evaluation.}
\citet{liu-etal-2023-evaluating}  and \citet{kamalloo2023hagrid} employ human evaluation methods in which annotators are tasked with assessing the variability of responses generated by either search engines or advanced language models equipped with additional retriever modules, adhering to the criteria outlined in \citep{rashkin2021measuring}.
Both studies corroborate that these systems often fall short in delivering statements backed with credible support and accurate citations.
To avoid the time-consuming human evaluation, \citet{gao-etal-2023-enabling, bohnet2022attributed} employ NLI models to provide automated AIS (AutoAIS).
\citet{yue-etal-2023-automatic} propose to repurpose related tasks to train a novel automatic evaluation model as well as prompting LMs as attribution evaluators.
Our proposed benchmark systematically evaluates the performance of these approaches under various settings and reveals the challenge of automatic attribution evaluation via in-depth error analysis.

%% file: latex/conclusion.tex
\section{Conclusion}\
In this paper, we propose \texttt{AttributionBench}, a comprehensive benchmark for assessing automatic attribution evaluators. We conduct extensive experiments and analysis with cutting-edge LLMs like fine-tuned GPT-3.5, GPT-4, and a variety of open-source LLMs.
Results systematically show that automatically evaluating attribution is still very challenging.
Through meticulous error analysis, we identify the principal obstacles in this endeavor, notably (1) the models' lack of sensitivity to detailed, fine-grained information and (2) the discrepancy between the information accessible by the model and that by human annotators. We hope our research will shed light on these issues and guide future efforts toward the development of more accurate and reliable automatic attribution evaluators.

%% file: latex/Limitations.tex
\section{Limitations}
Despite the comprehensive nature of our work in developing AttributionBench and conducting extensive experiments to understand the challenges of automatic attribution evaluation, our study has several limitations that warrant future investigation:

\paragraph{Dataset Diversity and Representativeness.} Although we meticulously sampled data from 7 different datasets covering various domains, our benchmark may still not capture the full spectrum of real-world scenarios that LLMs encounter.
We will continuously update our benchmark to align with the rapid developments in LLMs and changes in web content, as well as involving more detailed classification tasks in addition to binary classification.

\paragraph{Generalizability of Findings.} While our benchmark provides valuable insights into the challenges of automatic attribution evaluation, the generalizability of our findings may be limited by the specific datasets and evaluation setup we adopted. However, these findings give a more direct comparison and comprehensive understanding of current attribution evaluation systems. This could further inspire future efforts to build more robust attributed LLMs and attribution evaluators.

\paragraph{The coverage of attribution evaluation models.} Our study takes initial steps to systematically evaluate a diverse set of advanced attribution evaluators. There might still be missing ones which possess good performance on our benchmark, as it's impossible to cover all the existing models. Nonetheless, we hope our findings can shed light on building better attribution evaluators in the future.

%% file: latex/Ethics_Statement.tex

%% file: latex/appendix.tex
\clearpage
\appendix

\section{Implementation Details}
\label{sec:appendix_implementation_details}

\subsection{More Info about Source Datasets}
We introduce each source dataset used in \texttt{AttributionBench} and list the statistics in Appendix Table \ref{tab:appendix_data_statistics_complete}.
\paragraph{ExpertQA \citep{malaviya2023expertqa}} ExpertQA is constructed by collecting expert-curated questions from 484 participants spanning 32 fields of study. The same experts are then engaged to evaluate the generated responses to their own questions, considering both informativeness and attributionality.

\paragraph{Stanford-GenSearch \citep{liu-etal-2023-evaluating}}
Stanford-GenSearch utilizes statements extracted from diverse sources, including ELI5 \citep{eli5}, NaturalQuestions \citep{nq}, and AllSouls examinations, among others. These statements are input into popular and competent search engines to obtain responses. Human annotators then evaluate the precision and recall of the citations based on these responses.

\paragraph{AttributedQA \citep{rashkin2021measuring}}
This dataset comprises questions selected from the validation sets of Natural Questions \citep{nq}. Human evaluators are tasked with determining their attribution in adherence to the principles outlined in AIS \citep{rashkin2021measuring}.

\paragraph{LFQA \citep{chen2023understanding}}
LFQA consists of annotations for 100 questions randomly selected from the ELI-5 test set. Annotations were collected from six different language models (LM) paired with various document sets.

\paragraph{BEGIN \citep{dziri-etal-2022-evaluating}}
The BEGIN benchmark evaluates response attribution in dialogue systems, focusing on grounding responses to provided background knowledge. It comprises dialogue turns generated by advanced dialogue systems trained on three knowledge-grounded dialogue benchmarks and lets human annotators attribute responses to the background knowledge.

\paragraph{AttrEval-GenSearch \citep{yue-etal-2023-automatic}} This dataset is constructed by instructing annotators to formulate questions across 12 diverse domains and obtaining generated responses with citations from New Bing under balanced mode. To facilitate annotation, only the first sentence accompanied by a reference in the generation is considered.

\paragraph{HAGRID \citep{kamalloo2023hagrid}}
HAGRID is a dataset curated through a joint effort between LLMs and human evaluators, with queries sourced from MIRACL \citep{miracl}. The process involves obtaining responses from various LLMs and subsequently enlisting human evaluators to assess their verifiability.

\subsection{Label Processing}
We process each dataset into claim-level binary classification data. For the label space, we consider the example as ``attributable'' if the original label is in (``Attributable'', ``Fully attributable'' or ``Completely supported''), and convert all other labels such as ``Partially support'' into ``Not attributable''.

\subsection{Models}
\paragraph{Zero-shot prompting.} For GPT-3.5 and GPT-4, we use OpenAI's official APIs (\texttt{gpt-3.5-turbo-1106}, \texttt{gpt-4-1106-preview}). \footnote{\href{https://platform.openai.com/docs/api-reference/chat.}{https://platform.openai.com/docs/api-reference/chat.}} For other models' inference, the inputs are tokenized and truncated at a maximum of 2048 tokens. We generate text with a temperature of 0. The prompts for the task of evaluating attribution are provided in Appendix Table \ref{tab:appendix_prompts}.
\paragraph{Fine-tuning} For fine-tuning GPT-3.5, we use OpenAI's official API. \footnote{\href{https://platform.openai.com/finetune}{https://platform.openai.com/finetune}.
}
For fine-tuning other models, our implementation is based on the Huggingface library \citep{wolf-etal-2020-transformers}. The training is performed on 8 A100 80GB GPUs with a maximum of 2048 tokens. We use a batch size of 32 and train 2 epochs for all the models. We set the learning rate as 2e-5 and use a cosine learning rate decay with 0.03 warm-up steps. We used BF16, TF32, and FSDP \citep{10.14778/3611540.3611569} to efficiently train the models.

\input{tables/appendix_data_statistics_complete}

\input{figures/appendix_subset_errors}

\input{figures/appendix_detailed_error_categories}

\input{tables/appendix_prompts}

\input{tables/appendix_prompts_gpt}

\input{tables/appendix_prompts_gpt_2}

\section{Error Distribution in Sub-Datasets}
\label{sec:appendix_error_analysis}
We show the detailed error distribution in each sub-dataset in Figure \ref{fig:appendix_subset_error_cases} and Figure \ref{fig:appendix_detailed_error_categories}. Although ``fine-grained information insensitivity'' is the major error type among all datasets, the distribution of the rest error types still varies among 7 test sets. Moreover, the detailed error reasons within ``fine-grained information insensitivity'' are also diverse, indicating the diversity and difference among different test sets.

%% file: tables/appendix_data_statistics_complete.tex
\begin{table*}[t]
\resizebox{\linewidth}{!}{
\centering

\begin{tabular}{
p{0.15\linewidth}<{\raggedright}
p{0.2\linewidth}<{\raggedright}
p{0.2\linewidth}<{\raggedright}
p{0.2\linewidth}<{\raggedright}
p{0.12\linewidth}<{\raggedright}
p{0.12\linewidth}<{\raggedright}
p{0.12\linewidth}<{\raggedright}
p{0.08\linewidth}<{\raggedright}
p{0.08\linewidth}<{\raggedright}
}

\toprule[1pt]
\textbf{Dataset} & \textbf{\makecell[l]{Question\\Sources}} & \textbf{\makecell[l]{Response\\Sources}} & \textbf{\makecell[l]{Evidence\\Sources}} & \textbf{\makecell[l]{Claim\\Avg. Len.}} & \textbf{\makecell[l]{Evidence\\Avg. Len.}} & \textbf{\#Train} & \textbf{\#Dev} & \textbf{\#Test} \\
\midrule
ExpertQA & Curated by domain experts & GPT-4, Bing Chat & Google search results, Sphere & 34.0 & 515.0 & 4442 & 470 & 612 \\
\cmidrule{1-9}
Stanford-GenSearch & AllSouls, davinci-debate, ELI5, WikiHow-Keywords, NaturalQuestions & Bing Chat, NeevaAI, perplexity.ai, YouChat & Bing Chat, NeevaAI, perplexity.ai, YouChat & 25.1 & 465.2 & 1510 & 68 & 148 \\
\cmidrule{1-9}
AttributedQA & NaturalQuestions & PaLM (converted into long sentence with ChatGPT) & Wikipedia & 16.3 & 139.5 & 2000 & 74 & 230 \\
\cmidrule{1-9}
LFQA & ELI5 & WebGPT, GPT-3.5, Alpaca & WebGPT docs, Human docs & 26.3 & 141.4 & 2096 & 110 & 168 \\
\cmidrule{1-9}
\textbf{ID-Total} & & & & & & 13322 & 1198 & 1610 \\
\midrule[1pt]
BEGIN & Wizard-of-Wikipedia, CMU-DoG, TopicalChat & GPT2-base, T5-base, CTRL-\textsc{Dialog}, DoHA & Wikipedia, Reddit, news articles & 17.3 & 181.2 & \text{-} & \text{-} & 436 \\
\cmidrule{1-9}
HAGRID & MIRACL & GPT-3.5 & Wikipedia & 35.3 & 148.8 & \text{-} & \text{-} & 1088 \\
\cmidrule{1-9}
AttrEval-GenSearch & Curated by human annotators & New Bing & New Bing & 34.3 & 76.3 & \text{-} & \text{-} & 162 \\
\cmidrule{1-9}
\textbf{OOD-Total} & & & & & & \text{-} & \text{-} & 1686 \\

\bottomrule[1pt]
\end{tabular}
}
\caption{Statistics of the datasets used in \texttt{AttributionBench}.
They contain a wide range of diverse questions. ``ID'' and ``OOD'' denote "in-distribution" and "out-of-distribution", respectively. We make all subsets label-balanced
so that the evaluation can fairly show the performance of each class.
}
\label{tab:appendix_data_statistics_complete}
\end{table*}

%% file: figures/appendix_subset_errors.tex
\begin{figure*}[t]
    \centering
    \begin{minipage}{\textwidth}
        \includegraphics[width=\textwidth]{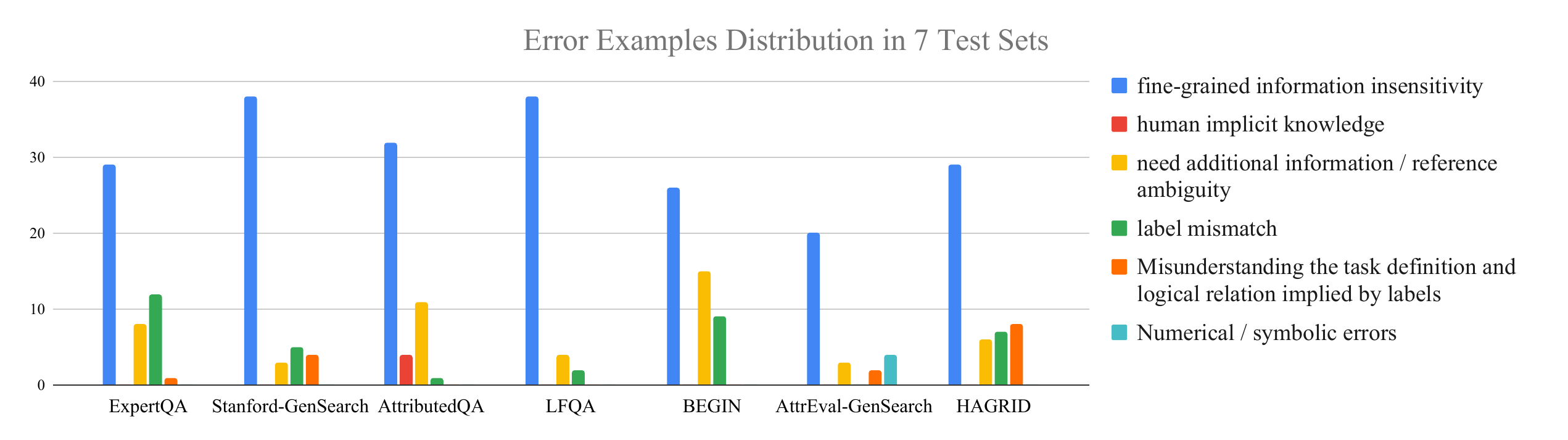}
        \caption{The distribution of error types among 7 test sets.}
        \label{fig:appendix_subset_error_cases}
    \end{minipage}
\end{figure*}

%% file: figures/appendix_detailed_error_categories.tex
\begin{figure*}[t]
    \centering
    \begin{minipage}{\textwidth}
        \includegraphics[width=\textwidth]{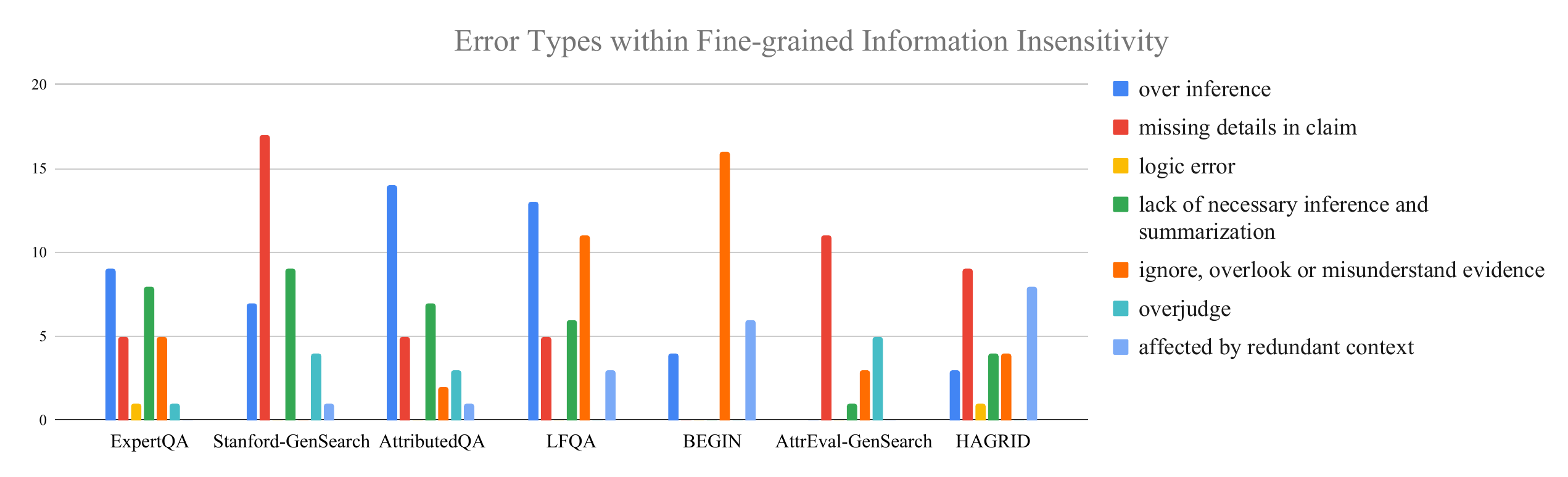}
        \caption{The distribution of error types within the major error reason ``fine-grained information insensitivity'' among 7 test sets.}
        \label{fig:appendix_detailed_error_categories}
    \end{minipage}
\end{figure*}

%% file: tables/appendix_prompts.tex
\begin{table*}[t]
\resizebox{\linewidth}{!}{
\small
\centering
\begin{tabularx}{\linewidth}{p{0.2\textwidth}<{\raggedright}l}%

\toprule[1pt]
\textbf{Prompt Name} & \textbf{Prompt Content} \\ \midrule
base\_c\_e & \makecell[Xt]{\#\#\# Instruction: \\ Please verify whether the reference supports the claim. Options: 'attributable' or 'not attributable'. \\\\ \#\#\#Input: \\ Claim: [Answer] \textbackslash{n}\textbackslash{n} \\ References: [concatenation of cited evidence] \\\\ \#\#\# Output:} \\
\midrule
base\_q\_c\_e & \makecell[Xt]{\#\#\# Instruction: \\ Please verify whether the reference supports the claim. The question is provided to help you better understand the claim, but your judgment should and only should be based on the claim, without being affected by the question. Options: 'attributable' or 'not attributable'.\\\\
\#\#\#Input: \\
\textcolor{teal}{Question: [Question]\textbackslash{n}\textbackslash{n}}\\
Claim: [Answer]\textbackslash{n}\textbackslash{n}\\
References: [concatenation of cited evidence] \\\\ \#\#\# Output:} \\
\midrule
base\_c\_e\_r & \makecell[Xt]{\#\#\# Instruction: \\ Please verify whether the reference supports the claim. The claim may be part of a response, or it may be the same as the response. The response is provided to help you better understand the claim, but your judgment should and only should be based on the claim, without being affected by the response. Options: 'attributable' or 'not attributable'.\\\\
\#\#\#Input:\\
Claim: [Answer] \textbackslash{n}\textbackslash{n}\\
\textcolor{teal}{Response: [Response]\textbackslash{n}\textbackslash{n}}\\
References: [concatenation of cited evidence]\\\\
\#\#\# Output:} \\
\midrule
base\_q\_c\_e\_r & \makecell[Xt]{\#\#\# Instruction: \\ Please verify whether the reference supports the claim. The claim may be part of a response, or it may be the same as the response. The question and the response are provided to help you better understand the claim, but your judgment should and only should be based on the claim, without being affected by the response or the question. Options: 'attributable' or 'not attributable'.
\\\\
\#\#\#Input:\\
\textcolor{teal}{Question: [Question]\textbackslash{n}\textbackslash{n}}\\
Claim: [Answer] \textbackslash{n}\textbackslash{n}\\
\textcolor{teal}{Response: [Response]\textbackslash{n}\textbackslash{n}}\\
References: [concatenation of cited evidence]\\\\
\#\#\# Output:} \\
\bottomrule[1pt]
\end{tabularx}
}
\caption{Prompts for involving different input fields into the model input. ``q'', ``c'', ``e'', and ``r'' represent question, claim, evidence, and response, respectively. The \textcolor{teal}{teal text} means the fields in addition to the base\_c\_e prompt. We use the ``base\_c\_e'' prompt for our main experiments as default since it achieves the best performance among all settings.}
\label{tab:appendix_prompts}
\end{table*}

%% file: tables/appendix_prompts_gpt.tex
\begin{table*}[t]
\resizebox{\linewidth}{!}{
\small
\centering
\begin{tabularx}{\linewidth}{p{0.2\textwidth}<{\raggedright}l}%

\toprule[1pt]
\textbf{Prompt Name} & \textbf{Prompt Content} \\ \midrule
base\_c\_e & \makecell[Xt]{\#\#\# Instruction: \\ Please verify whether the reference supports the claim. Options: 'attributable' or 'not attributable'. \\\\ \#\#\#Input: \\ Claim: [Answer] \textbackslash{n}\textbackslash{n} \\ References: [concatenation of cited evidence] \\\\ \#\#\# Output:} \\
\midrule
base\_c\_e (w/ CoT) & \makecell[Xt]{\#\#\# Instruction: \\ Please verify whether the reference supports the claim. \textcolor{blue}{Please first give your explanation, and then output your final judgment label} (Options: 'attributable' or 'not attributable'). \\\\ \#\#\#Input: \\ Claim: [Answer] \textbackslash{n}\textbackslash{n} \\ References: [concatenation of cited evidence] \\\\ \#\#\# Output:} \\
\midrule
breakdown\_c\_e (w/ CoT) & \makecell[Xt]{
\#\#\# Instruction:\\
\textcolor{blue}{Your task is to determine if a given claim is directly and explicitly supported by a provided reference. This involves a binary decision: labeling the claim as either 'attributable' or 'not attributable'.}\\
\textcolor{blue}{\#\#\# Procedure:}\\
\textcolor{blue}{Break Down the Claim and Reference: Carefully dissect both the claim and the reference into their fundamental components. This will help in pinpointing the exact areas of alignment or misalignment.}\\
Please first give your explanation, and then output your final judgment label (Options: 'attributable' or 'not attributable').\\\\
\#\#\#Input: \\
Claim: [Answer] \textbackslash{n}\textbackslash{n}\\
References: [concatenation of cited evidence]\\\\
\#\#\# Output:} \\
\bottomrule[1pt]
\end{tabularx}
}
\caption{Prompt variations for testing the impact of different instructions on the results. We control the variable and add one description at a time. The \textcolor{blue}{blue text} shows the additional content added based on the previous prompt.}
\label{tab:appendix_prompts_gpt}
\end{table*}

%% file: tables/appendix_prompts_gpt_2.tex
\begin{table*}[t]
\resizebox{0.952\linewidth}{!}{
\small
\centering
\begin{tabularx}{\linewidth}{p{0.2\textwidth}<{\raggedright}l}%

\toprule[1pt]
\textbf{Prompt Name} & \textbf{Prompt Content} \\ \midrule
\makecell[Xt]{breakdown\_c\_e (w/ CoT)\\+detail\_check} & \makecell[Xt]{
\#\#\# Instruction:\\
Your task is to determine if a given claim is directly and explicitly supported by a provided reference. This involves a binary decision: labeling the claim as either 'attributable' or 'not attributable'.\\
\#\#\# Procedure:\\
Break Down the Claim and Reference: Carefully dissect both the claim and the reference into their fundamental components. This will help in pinpointing the exact areas of alignment or misalignment.\\
\textcolor{blue}{Direct and Explicit Support:}\\
\textcolor{blue}{Attributable: Label the claim as 'attributable' if each component of the claim is directly and explicitly supported by the reference. Look for clear, specific mentions or unambiguous implications in the reference that align with the claim.}\\
\textcolor{blue}{Not Attributable: If any component of the claim lacks this direct and explicit support in the reference, label the claim as 'not attributable'.}\\
\textcolor{blue}{Avoid Over-Interpretation:}\\
\textcolor{blue}{Be cautious of inferring too much from the reference. If the support for a component of the claim requires significant interpretation or assumption beyond what is clearly stated, it is likely a case for 'not attributable'}.\\
Please first give your explanation, and then output your final judgment label (Options: 'attributable' or 'not attributable').\\\\
\#\#\#Input: \\
Claim: [Answer] \textbackslash{n}\textbackslash{n}\\
References: [concatenation of cited evidence]\\\\
\#\#\# Output:} \\

\midrule

\makecell[Xt]{breakdown\_c\_e (w/ CoT)\\+detail\_check\\+allow\_inference} & \makecell[Xt]{\#\#\# Instruction:\\
Your task is to determine if a given claim is directly and explicitly supported by a provided reference. This involves a binary decision: labeling the claim as either 'attributable' or 'not attributable'.\\
\#\#\# Procedure:\\
Break Down the Claim and Reference: Carefully dissect both the claim and the reference into their fundamental components. This will help in pinpointing the exact areas of alignment or misalignment.\\
Direct and Explicit Support:\\
Attributable: Label the claim as 'attributable' if each component of the claim is directly and explicitly supported by the reference. Look for clear, specific mentions or unambiguous implications in the reference that align with the claim.\\
Not Attributable: If any component of the claim lacks this direct and explicit support in the reference, label the claim as 'not attributable'.\\
Avoid Over-Interpretation:\\
Be cautious of inferring too much from the reference. If the support for a component of the claim requires significant interpretation or assumption beyond what is clearly stated, it is likely a case for 'not attributable'.\\
\textcolor{blue}{Comprehensive Understanding:}\\
\textcolor{blue}{The goal is not to align every aspect of factual information within the claim and the reference, since the reference could be very long, and there might be some redundant information. However, everything that occurs in the claim, must be supported within the reference. Also, since the claim could be a substring of a long response, so there might be some pronouns that are not linked to any entities, but you can treat these as the topic in the reference, apparently. Furthermore, you should understand the semantics of both the claim and the response, understand what they are trying to express. After that, you can somehow summarize the reference and the claim, and do a little bit inference between them. When you check the factual information, focus on those are obviously contradict, and mark these as 'not attributable'. However, sometimes the narrative in the claim and the response might be a little bit vague but still somehow inferencable, then don't be too strict and judge them as attributable. If all factual information in the claim can be found in the reference, or if the reference provides a thematic context that supports the claim, judge it as 'attributable'. If there are spans in the claim that are not factually verified or the thematic context does not align with the claim, then output 'not attributable'.}\\
Please first give your explanation, and then output your final judgment label (Options: 'attributable' or 'not attributable').\\\\
\#\#\#Input: \\
Claim: [Answer] \textbackslash{n}\textbackslash{n}\\
References: [concatenation of cited evidence]\\\\
\#\#\# Output:} \\

\bottomrule[1pt]
\end{tabularx}
}
\caption{(Continue from Table \ref{tab:appendix_prompts_gpt}) Prompt variations for testing the impact of different instructions on the results. We control the variable and add one description at a time. The \textcolor{blue}{blue text} shows the additional content added based on the previous prompt.}
\label{tab:appendix_prompts_gpt_2}
\end{table*}

%% file: acl_latex.bbl
\begin{thebibliography}{40}
\expandafter\ifx\csname natexlab\endcsname\relax\def\natexlab#1{#1}\fi

\bibitem[{Achiam et~al.(2023)Achiam, Adler, Agarwal, Ahmad, Akkaya, Aleman, Almeida, Altenschmidt, Altman, Anadkat et~al.}]{achiam2023gpt}
Josh Achiam, Steven Adler, Sandhini Agarwal, Lama Ahmad, Ilge Akkaya, Florencia~Leoni Aleman, Diogo Almeida, Janko Altenschmidt, Sam Altman, Shyamal Anadkat, et~al. 2023.
\newblock \href {https://arxiv.org/pdf/2303.08774.pdf} {Gpt-4 technical report}.
\newblock \emph{arXiv preprint arXiv:2303.08774}.

\bibitem[{Bohnet et~al.(2022)Bohnet, Tran, Verga, Aharoni, Andor, Soares, Eisenstein, Ganchev, Herzig, Hui et~al.}]{bohnet2022attributed}
Bernd Bohnet, Vinh~Q Tran, Pat Verga, Roee Aharoni, Daniel Andor, Livio~Baldini Soares, Jacob Eisenstein, Kuzman Ganchev, Jonathan Herzig, Kai Hui, et~al. 2022.
\newblock \href {https://arxiv.org/abs/2212.08037} {Attributed question answering: Evaluation and modeling for attributed large language models}.
\newblock \emph{arXiv preprint arXiv:2212.08037}.

\bibitem[{Brown et~al.(2020)Brown, Mann, Ryder, Subbiah, Kaplan, Dhariwal, Neelakantan, Shyam, Sastry, Askell, Agarwal, Herbert{-}Voss, Krueger, Henighan, Child, Ramesh, Ziegler, Wu, Winter, Hesse, Chen, Sigler, Litwin, Gray, Chess, Clark, Berner, McCandlish, Radford, Sutskever, and Amodei}]{gpt3}
Tom~B. Brown, Benjamin Mann, Nick Ryder, Melanie Subbiah, Jared Kaplan, Prafulla Dhariwal, Arvind Neelakantan, Pranav Shyam, Girish Sastry, Amanda Askell, Sandhini Agarwal, Ariel Herbert{-}Voss, Gretchen Krueger, Tom Henighan, Rewon Child, Aditya Ramesh, Daniel~M. Ziegler, Jeffrey Wu, Clemens Winter, Christopher Hesse, Mark Chen, Eric Sigler, Mateusz Litwin, Scott Gray, Benjamin Chess, Jack Clark, Christopher Berner, Sam McCandlish, Alec Radford, Ilya Sutskever, and Dario Amodei. 2020.
\newblock \href {https://proceedings.neurips.cc/paper/2020/hash/1457c0d6bfcb4967418bfb8ac142f64a-Abstract.html} {Language models are few-shot learners}.
\newblock In \emph{Advances in Neural Information Processing Systems 33: Annual Conference on Neural Information Processing Systems 2020, NeurIPS 2020, December 6-12, 2020, virtual}.

\bibitem[{Chen et~al.(2023)Chen, Xu, Arora, and Choi}]{chen2023understanding}
Hung-Ting Chen, Fangyuan Xu, Shane~A Arora, and Eunsol Choi. 2023.
\newblock \href {https://arxiv.org/abs/2310.12150} {Understanding retrieval augmentation for long-form question answering}.
\newblock \emph{arXiv preprint arXiv:2310.12150}.

\bibitem[{Chowdhery et~al.(2023)Chowdhery, Narang, Devlin, Bosma, Mishra, Roberts, Barham, Chung, Sutton, Gehrmann et~al.}]{chowdhery2023palm}
Aakanksha Chowdhery, Sharan Narang, Jacob Devlin, Maarten Bosma, Gaurav Mishra, Adam Roberts, Paul Barham, Hyung~Won Chung, Charles Sutton, Sebastian Gehrmann, et~al. 2023.
\newblock \href {https://www.jmlr.org/papers/v24/22-1144.html} {Palm: Scaling language modeling with pathways}.
\newblock \emph{Journal of Machine Learning Research}, 24(240):1--113.

\bibitem[{Chung et~al.(2022)Chung, Hou, Longpre, Zoph, Tay, Fedus, Li, Wang, Dehghani, Brahma et~al.}]{chung2022scaling}
Hyung~Won Chung, Le~Hou, Shayne Longpre, Barret Zoph, Yi~Tay, William Fedus, Eric Li, Xuezhi Wang, Mostafa Dehghani, Siddhartha Brahma, et~al. 2022.
\newblock \href {https://arxiv.org/abs/2210.11416} {Scaling instruction-finetuned language models}.
\newblock \emph{arXiv preprint arXiv:2210.11416}.

\bibitem[{Dziri et~al.(2022{\natexlab{a}})Dziri, Milton, Yu, Zaiane, and Reddy}]{dziri-etal-2022-origin}
Nouha Dziri, Sivan Milton, Mo~Yu, Osmar Zaiane, and Siva Reddy. 2022{\natexlab{a}}.
\newblock \href {https://doi.org/10.18653/v1/2022.naacl-main.387} {On the origin of hallucinations in conversational models: Is it the datasets or the models?}
\newblock In \emph{Proceedings of the 2022 Conference of the North American Chapter of the Association for Computational Linguistics: Human Language Technologies}, pages 5271--5285, Seattle, United States. Association for Computational Linguistics.

\bibitem[{Dziri et~al.(2022{\natexlab{b}})Dziri, Rashkin, Linzen, and Reitter}]{dziri-etal-2022-evaluating}
Nouha Dziri, Hannah Rashkin, Tal Linzen, and David Reitter. 2022{\natexlab{b}}.
\newblock \href {https://doi.org/10.1162/tacl_a_00506} {Evaluating attribution in dialogue systems: The {BEGIN} benchmark}.
\newblock \emph{Transactions of the Association for Computational Linguistics}, 10:1066--1083.

\bibitem[{Fan et~al.(2019)Fan, Jernite, Perez, Grangier, Weston, and Auli}]{eli5}
Angela Fan, Yacine Jernite, Ethan Perez, David Grangier, Jason Weston, and Michael Auli. 2019.
\newblock \href {https://doi.org/10.18653/v1/P19-1346} {{ELI}5: Long form question answering}.
\newblock In \emph{Proceedings of the 57th Annual Meeting of the Association for Computational Linguistics}, pages 3558--3567, Florence, Italy. Association for Computational Linguistics.

\bibitem[{Gao et~al.(2023{\natexlab{a}})Gao, Dai, Pasupat, Chen, Chaganty, Fan, Zhao, Lao, Lee, Juan, and Guu}]{gao-etal-2023-rarr}
Luyu Gao, Zhuyun Dai, Panupong Pasupat, Anthony Chen, Arun~Tejasvi Chaganty, Yicheng Fan, Vincent Zhao, Ni~Lao, Hongrae Lee, Da-Cheng Juan, and Kelvin Guu. 2023{\natexlab{a}}.
\newblock \href {https://doi.org/10.18653/v1/2023.acl-long.910} {{RARR}: Researching and revising what language models say, using language models}.
\newblock In \emph{Proceedings of the 61st Annual Meeting of the Association for Computational Linguistics (Volume 1: Long Papers)}, pages 16477--16508, Toronto, Canada. Association for Computational Linguistics.

\bibitem[{Gao et~al.(2023{\natexlab{b}})Gao, Yen, Yu, and Chen}]{gao-etal-2023-enabling}
Tianyu Gao, Howard Yen, Jiatong Yu, and Danqi Chen. 2023{\natexlab{b}}.
\newblock \href {https://doi.org/10.18653/v1/2023.emnlp-main.398} {Enabling large language models to generate text with citations}.
\newblock In \emph{Proceedings of the 2023 Conference on Empirical Methods in Natural Language Processing}, pages 6465--6488, Singapore. Association for Computational Linguistics.

\bibitem[{Glaese et~al.(2022)Glaese, McAleese, Trębacz, Aslanides, Firoiu, Ewalds, Rauh, Weidinger, Chadwick, Thacker, Campbell-Gillingham, Uesato, Huang, Comanescu, Yang, See, Dathathri, Greig, Chen, Fritz, Elias, Green, Mokrá, Fernando, Wu, Foley, Young, Gabriel, Isaac, Mellor, Hassabis, Kavukcuoglu, Hendricks, and Irving}]{sparrow}
Amelia Glaese, Nat McAleese, Maja Trębacz, John Aslanides, Vlad Firoiu, Timo Ewalds, Maribeth Rauh, Laura Weidinger, Martin Chadwick, Phoebe Thacker, Lucy Campbell-Gillingham, Jonathan Uesato, Po-Sen Huang, Ramona Comanescu, Fan Yang, Abigail See, Sumanth Dathathri, Rory Greig, Charlie Chen, Doug Fritz, Jaume~Sanchez Elias, Richard Green, Soňa Mokrá, Nicholas Fernando, Boxi Wu, Rachel Foley, Susannah Young, Iason Gabriel, William Isaac, John Mellor, Demis Hassabis, Koray Kavukcuoglu, Lisa~Anne Hendricks, and Geoffrey Irving. 2022.
\newblock \href {https://arxiv.org/abs/2209.14375} {Improving alignment of dialogue agents via targeted human judgements}.
\newblock \emph{ArXiv preprint}, abs/2209.14375.

\bibitem[{Guu et~al.(2020)Guu, Lee, Tung, Pasupat, and Chang}]{pmlr-v119-guu20a}
Kelvin Guu, Kenton Lee, Zora Tung, Panupong Pasupat, and Mingwei Chang. 2020.
\newblock \href {https://proceedings.mlr.press/v119/guu20a.html} {Retrieval augmented language model pre-training}.
\newblock In \emph{Proceedings of the 37th International Conference on Machine Learning}, volume 119 of \emph{Proceedings of Machine Learning Research}, pages 3929--3938. PMLR.

\bibitem[{Honovich et~al.(2022)Honovich, Aharoni, Herzig, Taitelbaum, Kukliansy, Cohen, Scialom, Szpektor, Hassidim, and Matias}]{honovich-etal-2022-true-evaluating}
Or~Honovich, Roee Aharoni, Jonathan Herzig, Hagai Taitelbaum, Doron Kukliansy, Vered Cohen, Thomas Scialom, Idan Szpektor, Avinatan Hassidim, and Yossi Matias. 2022.
\newblock \href {https://doi.org/10.18653/v1/2022.naacl-main.287} {{TRUE}: Re-evaluating factual consistency evaluation}.
\newblock In \emph{Proceedings of the 2022 Conference of the North American Chapter of the Association for Computational Linguistics: Human Language Technologies}, pages 3905--3920, Seattle, United States. Association for Computational Linguistics.

\bibitem[{Ji et~al.(2023)Ji, Lee, Frieske, Yu, Su, Xu, Ishii, Bang, Madotto, and Fung}]{10.1145/3571730}
Ziwei Ji, Nayeon Lee, Rita Frieske, Tiezheng Yu, Dan Su, Yan Xu, Etsuko Ishii, Ye~Jin Bang, Andrea Madotto, and Pascale Fung. 2023.
\newblock \href {https://doi.org/10.1145/3571730} {Survey of hallucination in natural language generation}.
\newblock \emph{ACM Comput. Surv.}, 55(12).

\bibitem[{Kamalloo et~al.(2023)Kamalloo, Jafari, Zhang, Thakur, and Lin}]{kamalloo2023hagrid}
Ehsan Kamalloo, Aref Jafari, Xinyu Zhang, Nandan Thakur, and Jimmy Lin. 2023.
\newblock \href {https://arxiv.org/abs/2307.16883} {Hagrid: A human-llm collaborative dataset for generative information-seeking with attribution}.
\newblock \emph{ArXiv preprint}, abs/2307.16883.

\bibitem[{Komeili et~al.(2022)Komeili, Shuster, and Weston}]{komeili-etal-2022-internet}
Mojtaba Komeili, Kurt Shuster, and Jason Weston. 2022.
\newblock \href {https://doi.org/10.18653/v1/2022.acl-long.579} {{I}nternet-augmented dialogue generation}.
\newblock In \emph{Proceedings of the 60th Annual Meeting of the Association for Computational Linguistics (Volume 1: Long Papers)}, pages 8460--8478, Dublin, Ireland. Association for Computational Linguistics.

\bibitem[{Kwiatkowski et~al.(2019)Kwiatkowski, Palomaki, Redfield, Collins, Parikh, Alberti, Epstein, Polosukhin, Devlin, Lee, Toutanova, Jones, Kelcey, Chang, Dai, Uszkoreit, Le, and Petrov}]{nq}
Tom Kwiatkowski, Jennimaria Palomaki, Olivia Redfield, Michael Collins, Ankur Parikh, Chris Alberti, Danielle Epstein, Illia Polosukhin, Jacob Devlin, Kenton Lee, Kristina Toutanova, Llion Jones, Matthew Kelcey, Ming-Wei Chang, Andrew~M. Dai, Jakob Uszkoreit, Quoc Le, and Slav Petrov. 2019.
\newblock \href {https://doi.org/10.1162/tacl_a_00276} {Natural questions: A benchmark for question answering research}.
\newblock \emph{Transactions of the Association for Computational Linguistics}, 7:452--466.

\bibitem[{Lewis et~al.(2020)Lewis, Perez, Piktus, Petroni, Karpukhin, Goyal, K{\"{u}}ttler, Lewis, Yih, Rockt{\"{a}}schel, Riedel, and Kiela}]{2005.11401v4}
Patrick S.~H. Lewis, Ethan Perez, Aleksandra Piktus, Fabio Petroni, Vladimir Karpukhin, Naman Goyal, Heinrich K{\"{u}}ttler, Mike Lewis, Wen{-}tau Yih, Tim Rockt{\"{a}}schel, Sebastian Riedel, and Douwe Kiela. 2020.
\newblock \href {https://proceedings.neurips.cc/paper/2020/hash/6b493230205f780e1bc26945df7481e5-Abstract.html} {Retrieval-augmented generation for knowledge-intensive {NLP} tasks}.
\newblock In \emph{Advances in Neural Information Processing Systems 33: Annual Conference on Neural Information Processing Systems 2020, NeurIPS 2020, December 6-12, 2020, virtual}.

\bibitem[{Li et~al.(2023)Li, Zhao, Yu, Song, Li, Yu, Li, Huang, and Li}]{li-etal-2023-api}
Minghao Li, Yingxiu Zhao, Bowen Yu, Feifan Song, Hangyu Li, Haiyang Yu, Zhoujun Li, Fei Huang, and Yongbin Li. 2023.
\newblock \href {https://doi.org/10.18653/v1/2023.emnlp-main.187} {{API}-bank: A comprehensive benchmark for tool-augmented {LLM}s}.
\newblock In \emph{Proceedings of the 2023 Conference on Empirical Methods in Natural Language Processing}, pages 3102--3116, Singapore. Association for Computational Linguistics.

\bibitem[{Liu et~al.(2023)Liu, Zhang, and Liang}]{liu-etal-2023-evaluating}
Nelson Liu, Tianyi Zhang, and Percy Liang. 2023.
\newblock \href {https://doi.org/10.18653/v1/2023.findings-emnlp.467} {Evaluating verifiability in generative search engines}.
\newblock In \emph{Findings of the Association for Computational Linguistics: EMNLP 2023}, pages 7001--7025, Singapore. Association for Computational Linguistics.

\bibitem[{Liu et~al.(2019)Liu, Ott, Goyal, Du, Joshi, Chen, Levy, Lewis, Zettlemoyer, and Stoyanov}]{liu2019roberta}
Yinhan Liu, Myle Ott, Naman Goyal, Jingfei Du, Mandar Joshi, Danqi Chen, Omer Levy, Mike Lewis, Luke Zettlemoyer, and Veselin Stoyanov. 2019.
\newblock \href {https://arxiv.org/abs/1907.11692} {Roberta: A robustly optimized bert pretraining approach}.
\newblock \emph{arXiv preprint arXiv:1907.11692}.

\bibitem[{Malaviya et~al.(2023)Malaviya, Lee, Chen, Sieber, Yatskar, and Roth}]{malaviya2023expertqa}
Chaitanya Malaviya, Subin Lee, Sihao Chen, Elizabeth Sieber, Mark Yatskar, and Dan Roth. 2023.
\newblock \href {https://arxiv.org/abs/2309.07852} {Expertqa: Expert-curated questions and attributed answers}.
\newblock \emph{ArXiv preprint}, abs/2309.07852.

\bibitem[{Menick et~al.(2022)Menick, Trebacz, Mikulik, Aslanides, Song, Chadwick, Glaese, Young, Campbell-Gillingham, Irving, and McAleese}]{gophercite}
Jacob Menick, Maja Trebacz, Vladimir Mikulik, John Aslanides, Francis Song, Martin Chadwick, Mia Glaese, Susannah Young, Lucy Campbell-Gillingham, Geoffrey Irving, and Nat McAleese. 2022.
\newblock \href {https://arxiv.org/abs/2203.11147} {Teaching language models to support answers with verified quotes}.
\newblock \emph{ArXiv preprint}, abs/2203.11147.

\bibitem[{Mialon et~al.(2023)Mialon, Dessì, Lomeli, Nalmpantis, Pasunuru, Raileanu, Rozière, Schick, Dwivedi-Yu, Celikyilmaz, Grave, LeCun, and Scialom}]{2302.07842v1}
Grégoire Mialon, Roberto Dessì, Maria Lomeli, Christoforos Nalmpantis, Ram Pasunuru, Roberta Raileanu, Baptiste Rozière, Timo Schick, Jane Dwivedi-Yu, Asli Celikyilmaz, Edouard Grave, Yann LeCun, and Thomas Scialom. 2023.
\newblock \href {https://arxiv.org/abs/2302.07842} {Augmented language models: a survey}.
\newblock \emph{ArXiv preprint}, abs/2302.07842.

\bibitem[{Nakano et~al.(2021)Nakano, Hilton, Balaji, Wu, Ouyang, Kim, Hesse, Jain, Kosaraju, Saunders et~al.}]{nakano2021webgpt}
Reiichiro Nakano, Jacob Hilton, Suchir Balaji, Jeff Wu, Long Ouyang, Christina Kim, Christopher Hesse, Shantanu Jain, Vineet Kosaraju, William Saunders, et~al. 2021.
\newblock \href {https://arxiv.org/abs/2112.09332} {Webgpt: Browser-assisted question-answering with human feedback}.
\newblock \emph{arXiv preprint arXiv:2112.09332}.

\bibitem[{OpenAI(2023)}]{openai2023chatgpt}
OpenAI. 2023.
\newblock \href {https://chat.openai.com/chat} {Chatgpt (nov 06 version)}.
\newblock \emph{https://chat.openai.com/chat}.

\bibitem[{Ouyang et~al.(2022)Ouyang, Wu, Jiang, Almeida, Wainwright, Mishkin, Zhang, Agarwal, Slama, Ray et~al.}]{ouyang2022training}
Long Ouyang, Jeffrey Wu, Xu~Jiang, Diogo Almeida, Carroll Wainwright, Pamela Mishkin, Chong Zhang, Sandhini Agarwal, Katarina Slama, Alex Ray, et~al. 2022.
\newblock \href {https://proceedings.neurips.cc/paper_files/paper/2022/hash/b1efde53be364a73914f58805a001731-Abstract-Conference.html} {Training language models to follow instructions with human feedback}.
\newblock \emph{Advances in Neural Information Processing Systems}, 35:27730--27744.

\bibitem[{Peng et~al.(2023)Peng, Li, He, Galley, and Gao}]{peng2023instruction}
Baolin Peng, Chunyuan Li, Pengcheng He, Michel Galley, and Jianfeng Gao. 2023.
\newblock \href {https://arxiv.org/abs/2304.03277} {Instruction tuning with gpt-4}.
\newblock \emph{arXiv preprint arXiv:2304.03277}.

\bibitem[{Rashkin et~al.(2021)Rashkin, Nikolaev, Lamm, Aroyo, Collins, Das, Petrov, Tomar, Turc, and Reitter}]{rashkin2021measuring}
Hannah Rashkin, Vitaly Nikolaev, Matthew Lamm, Lora Aroyo, Michael Collins, Dipanjan Das, Slav Petrov, Gaurav~Singh Tomar, Iulia Turc, and David Reitter. 2021.
\newblock \href {https://arxiv.org/abs/2112.12870} {Measuring attribution in natural language generation models}.
\newblock \emph{arXiv preprint arXiv:2112.12870}.

\bibitem[{Schick et~al.(2023)Schick, Dwivedi-Yu, Dessì, Raileanu, Lomeli, Zettlemoyer, Cancedda, and Scialom}]{2302.04761v1}
Timo Schick, Jane Dwivedi-Yu, Roberto Dessì, Roberta Raileanu, Maria Lomeli, Luke Zettlemoyer, Nicola Cancedda, and Thomas Scialom. 2023.
\newblock \href {https://arxiv.org/abs/2302.04761} {Toolformer: Language models can teach themselves to use tools}.
\newblock \emph{ArXiv preprint}, abs/2302.04761.

\bibitem[{Tay et~al.(2022)Tay, Dehghani, Tran, Garcia, Wei, Wang, Chung, Bahri, Schuster, Zheng et~al.}]{tay2022ul2}
Yi~Tay, Mostafa Dehghani, Vinh~Q Tran, Xavier Garcia, Jason Wei, Xuezhi Wang, Hyung~Won Chung, Dara Bahri, Tal Schuster, Steven Zheng, et~al. 2022.
\newblock \href {https://arxiv.org/abs/2205.05131} {Ul2: Unifying language learning paradigms}.
\newblock In \emph{The Eleventh International Conference on Learning Representations}.

\bibitem[{Thoppilan et~al.(2022)Thoppilan, Freitas, Hall, Shazeer, Kulshreshtha, Cheng, Jin, Bos, Baker, Du, Li, Lee, Zheng, Ghafouri, Menegali, Huang, Krikun, Lepikhin, Qin, Chen, Xu, Chen, Roberts, Bosma, Zhao, Zhou, Chang, Krivokon, Rusch, Pickett, Srinivasan, Man, Meier-Hellstern, Morris, Doshi, Santos, Duke, Soraker, Zevenbergen, Prabhakaran, Diaz, Hutchinson, Olson, Molina, Hoffman-John, Lee, Aroyo, Rajakumar, Butryna, Lamm, Kuzmina, Fenton, Cohen, Bernstein, Kurzweil, Aguera-Arcas, Cui, Croak, Chi, and Le}]{lamda}
Romal Thoppilan, Daniel~De Freitas, Jamie Hall, Noam Shazeer, Apoorv Kulshreshtha, Heng-Tze Cheng, Alicia Jin, Taylor Bos, Leslie Baker, Yu~Du, YaGuang Li, Hongrae Lee, Huaixiu~Steven Zheng, Amin Ghafouri, Marcelo Menegali, Yanping Huang, Maxim Krikun, Dmitry Lepikhin, James Qin, Dehao Chen, Yuanzhong Xu, Zhifeng Chen, Adam Roberts, Maarten Bosma, Vincent Zhao, Yanqi Zhou, Chung-Ching Chang, Igor Krivokon, Will Rusch, Marc Pickett, Pranesh Srinivasan, Laichee Man, Kathleen Meier-Hellstern, Meredith~Ringel Morris, Tulsee Doshi, Renelito~Delos Santos, Toju Duke, Johnny Soraker, Ben Zevenbergen, Vinodkumar Prabhakaran, Mark Diaz, Ben Hutchinson, Kristen Olson, Alejandra Molina, Erin Hoffman-John, Josh Lee, Lora Aroyo, Ravi Rajakumar, Alena Butryna, Matthew Lamm, Viktoriya Kuzmina, Joe Fenton, Aaron Cohen, Rachel Bernstein, Ray Kurzweil, Blaise Aguera-Arcas, Claire Cui, Marian Croak, Ed~Chi, and Quoc Le. 2022.
\newblock \href {https://arxiv.org/abs/2201.08239} {Lamda: Language models for dialog applications}.
\newblock \emph{ArXiv preprint}, abs/2201.08239.

\bibitem[{Touvron et~al.(2023)Touvron, Martin, Stone, Albert, Almahairi, Babaei, Bashlykov, Batra, Bhargava, Bhosale et~al.}]{touvron2023llama}
Hugo Touvron, Louis Martin, Kevin Stone, Peter Albert, Amjad Almahairi, Yasmine Babaei, Nikolay Bashlykov, Soumya Batra, Prajjwal Bhargava, Shruti Bhosale, et~al. 2023.
\newblock \href {https://arxiv.org/abs/2302.13971} {Llama 2: Open foundation and fine-tuned chat models}.
\newblock \emph{arXiv preprint arXiv:2307.09288}.

\bibitem[{Wei et~al.(2021)Wei, Bosma, Zhao, Guu, Yu, Lester, Du, Dai, and Le}]{wei2021finetuned}
Jason Wei, Maarten Bosma, Vincent Zhao, Kelvin Guu, Adams~Wei Yu, Brian Lester, Nan Du, Andrew~M Dai, and Quoc~V Le. 2021.
\newblock \href {https://openreview.net/forum?id=gEZrGCozdqR} {Finetuned language models are zero-shot learners}.
\newblock In \emph{International Conference on Learning Representations}.

\bibitem[{Wei et~al.(2022)Wei, Wang, Schuurmans, Bosma, Xia, Chi, Le, Zhou et~al.}]{wei2022chain}
Jason Wei, Xuezhi Wang, Dale Schuurmans, Maarten Bosma, Fei Xia, Ed~Chi, Quoc~V Le, Denny Zhou, et~al. 2022.
\newblock \href {https://proceedings.neurips.cc/paper_files/paper/2022/hash/9d5609613524ecf4f15af0f7b31abca4-Abstract-Conference.html} {Chain-of-thought prompting elicits reasoning in large language models}.
\newblock \emph{Advances in Neural Information Processing Systems}, 35:24824--24837.

\bibitem[{Wolf et~al.(2020)Wolf, Debut, Sanh, Chaumond, Delangue, Moi, Cistac, Rault, Louf, Funtowicz, Davison, Shleifer, von Platen, Ma, Jernite, Plu, Xu, Le~Scao, Gugger, Drame, Lhoest, and Rush}]{wolf-etal-2020-transformers}
Thomas Wolf, Lysandre Debut, Victor Sanh, Julien Chaumond, Clement Delangue, Anthony Moi, Pierric Cistac, Tim Rault, Remi Louf, Morgan Funtowicz, Joe Davison, Sam Shleifer, Patrick von Platen, Clara Ma, Yacine Jernite, Julien Plu, Canwen Xu, Teven Le~Scao, Sylvain Gugger, Mariama Drame, Quentin Lhoest, and Alexander Rush. 2020.
\newblock \href {https://doi.org/10.18653/v1/2020.emnlp-demos.6} {Transformers: State-of-the-art natural language processing}.
\newblock In \emph{Proceedings of the 2020 Conference on Empirical Methods in Natural Language Processing: System Demonstrations}, pages 38--45, Online. Association for Computational Linguistics.

\bibitem[{Yue et~al.(2023)Yue, Wang, Chen, Zhang, Su, and Sun}]{yue-etal-2023-automatic}
Xiang Yue, Boshi Wang, Ziru Chen, Kai Zhang, Yu~Su, and Huan Sun. 2023.
\newblock \href {https://doi.org/10.18653/v1/2023.findings-emnlp.307} {Automatic evaluation of attribution by large language models}.
\newblock In \emph{Findings of the Association for Computational Linguistics: EMNLP 2023}, pages 4615--4635, Singapore. Association for Computational Linguistics.

\bibitem[{Zhang et~al.(2022)Zhang, Thakur, Ogundepo, Kamalloo, Alfonso-Hermelo, Li, Liu, Rezagholizadeh, and Lin}]{miracl}
Xinyu Zhang, Nandan Thakur, Odunayo Ogundepo, Ehsan Kamalloo, David Alfonso-Hermelo, Xiaoguang Li, Qun Liu, Mehdi Rezagholizadeh, and Jimmy Lin. 2022.
\newblock \href {https://arxiv.org/abs/2210.09984} {Making a miracl: Multilingual information retrieval across a continuum of languages}.
\newblock \emph{ArXiv preprint}, abs/2210.09984.

\bibitem[{Zhao et~al.(2023)Zhao, Gu, Varma, Luo, Huang, Xu, Wright, Shojanazeri, Ott, Shleifer, Desmaison, Balioglu, Damania, Nguyen, Chauhan, Hao, Mathews, and Li}]{10.14778/3611540.3611569}
Yanli Zhao, Andrew Gu, Rohan Varma, Liang Luo, Chien-Chin Huang, Min Xu, Less Wright, Hamid Shojanazeri, Myle Ott, Sam Shleifer, Alban Desmaison, Can Balioglu, Pritam Damania, Bernard Nguyen, Geeta Chauhan, Yuchen Hao, Ajit Mathews, and Shen Li. 2023.
\newblock \href {https://doi.org/10.14778/3611540.3611569} {Pytorch fsdp: Experiences on scaling fully sharded data parallel}.
\newblock \emph{Proc. VLDB Endow.}, 16(12):3848–3860.

\end{thebibliography}
